\RequirePackage[svgnames]{xcolor}

\documentclass[11pt,a4paper]{article}

\usepackage{todonotes}

\usepackage{etex} 

\usepackage{sidecap}
\usepackage{epsfig}
\usepackage{color}
\usepackage{multirow}
\usepackage{soul}
\usepackage{fancyhdr}
\usepackage{amsmath}
\usepackage{amssymb}
\usepackage{tcolorbox}
\usepackage{booktabs}
\usepackage{multirow}
\usepackage{longtable}

\definecolor{darknavy}{RGB}{0, 0, 102}

\usepackage{morewrites}
\definecolor{lightgreen}{rgb}{0.88,1,0.88}
\definecolor{lightred}{rgb}{1,0.88,0.88}
\definecolor{lightblue}{rgb}{0.88,0.93,1}
\definecolor{lightyellow}{rgb}{1,1,0.8}

\usepackage{pgfplots}
\usepackage{tikz}
\usetikzlibrary{positioning, arrows.meta}
\usepackage{longtable}
\usepackage{booktabs}
\usepackage{array}
\usepackage{adjustbox}
\usepackage{longtable}
\usepackage{pifont}
\usepackage{fontawesome5}
\usepackage{caption} 
\usepackage[justification=justified,singlelinecheck=false]{caption}
\usepackage{subcaption}
\usepackage{subfig}
\definecolor{indigo}{RGB}{75,0,130}
\definecolor{navy}{RGB}{0,0,128}
\definecolor{teal}{RGB}{0,128,128}

\newtcolorbox{defin}{colback=Teal!5!White,enhanced,title=Research Questions and Epistemic Vision of - \faDna \hspace{0.1cm} Neural Genomics,	attach boxed title to top left={xshift=-4mm},boxrule=0pt,after skip=1cm,before skip=1cm,right skip=0cm,breakable,fonttitle=\bfseries,toprule=0pt,bottomrule=0pt,rightrule=0pt,leftrule=3pt,arc=0mm,skin=enhancedlast jigsaw,sharp corners,colframe=Teal!55!black,colbacktitle=Teal!55!black,boxed title style={
		frame code={ 
			\fill[Teal!25!black](frame.south west)--(frame.north west)--(frame.north east)--([xshift=3mm]frame.east)--(frame.south east)--cycle;
			\draw[line width=1mm,Teal!25!black]([xshift=2mm]frame.north east)--([xshift=5mm]frame.east)--([xshift=2mm]frame.south east);
			\draw[line width=1mm,Teal!25!black]([xshift=5mm]frame.north east)--([xshift=8mm]frame.east)--([xshift=5mm]frame.south east);
			\fill[Teal!25!black](frame.south west)--+(4mm,-2mm)--+(4mm,2mm)--cycle;
		}
	}
}
\usetikzlibrary{shapes.geometric, arrows}
\usetikzlibrary{decorations.markings}

\newtcolorbox{defin_conclusion}{colback=Teal!5!White,enhanced,title=\faDna \hspace{0.1cm} Neural Genomics: A Paradigm to Comprehend the Semantic Organism of AI,	attach boxed title to top left={xshift=-4mm},boxrule=0pt,after skip=1cm,before skip=1cm,right skip=0cm,breakable,fonttitle=\bfseries,toprule=0pt,bottomrule=0pt,rightrule=0pt,leftrule=3pt,arc=0mm,skin=enhancedlast jigsaw,sharp corners,colframe=Teal!55!black,colbacktitle=Teal!55!black,boxed title style={
		frame code={ 
			\fill[Teal!25!black](frame.south west)--(frame.north west)--(frame.north east)--([xshift=3mm]frame.east)--(frame.south east)--cycle;
			\draw[line width=1mm,Teal!25!black]([xshift=2mm]frame.north east)--([xshift=5mm]frame.east)--([xshift=2mm]frame.south east);
			\draw[line width=1mm,Teal!25!black]([xshift=5mm]frame.north east)--([xshift=8mm]frame.east)--([xshift=5mm]frame.south east);
			\fill[Teal!25!black](frame.south west)--+(4mm,-2mm)--+(4mm,2mm)--cycle;
		}
	}
}
\definecolor{first}{RGB}{210,255,140}
\definecolor{second}{RGB}{136, 162, 190}
\definecolor{third}{RGB}{129, 222, 228}
\definecolor{fourth}{RGB}{132, 84, 246}
\definecolor{fifth}{RGB}{250, 223, 112}
\definecolor{sixth}{RGB}{203, 193, 172}
\definecolor{seventh}{RGB}{88, 112, 246}
\definecolor{eighth}{RGB}{245, 192, 106}
\definecolor{nine}{RGB}{171, 162, 111}
\definecolor{ten}{RGB}{217, 217, 217}

\definecolor{paired-light-blue}{RGB}{198, 219, 239}
\definecolor{paired-dark-blue}{RGB}{49, 130, 188}
\definecolor{paired-light-orange}{RGB}{251, 208, 162}
\definecolor{paired-dark-orange}{RGB}{230, 85, 12}
\definecolor{paired-light-green}{RGB}{199, 233, 193}
\definecolor{paired-dark-green}{RGB}{49, 163, 83}
\definecolor{paired-light-purple}{RGB}{218, 218, 235}
\definecolor{paired-dark-purple}{RGB}{117, 107, 176}
\definecolor{paired-light-gray}{RGB}{217, 217, 217}
\definecolor{paired-dark-gray}{RGB}{99, 99, 99}
\definecolor{paired-light-pink}{RGB}{222, 158, 214}
\definecolor{paired-dark-pink}{RGB}{123, 65, 115}
\definecolor{paired-light-red}{RGB}{231, 150, 156}
\definecolor{paired-dark-red}{RGB}{131, 60, 56}
\definecolor{paired-light-yellow}{RGB}{231, 204, 149}
\definecolor{paired-dark-yellow}{RGB}{141, 109, 49}
\definecolor{Teal}{RGB}{0, 50, 50}
\definecolor{White}{RGB}{250, 250, 250}

\definecolor{bg1}{HTML}{FF9966}
\definecolor{bg2}{HTML}{CCE5FF}
\definecolor{bg3}{HTML}{FFCC99}
\definecolor{bg4}{HTML}{FFC107}
\definecolor{bg5}{HTML}{FFCCCC}
\definecolor{bg6}{HTML}{D5E8D4}
\definecolor{bg7}{HTML}{eeeeee}
\definecolor{bg8}{HTML}{cdeb8b}
\definecolor{bg9}{HTML}{dae8fc}
\definecolor{bg10}{HTML}{a2e6eb}

\definecolor{bg31}{HTML}{FFCDD2} 
\definecolor{bg32}{HTML}{F8BBD0}
\definecolor{bg33}{HTML}{E1BEE7} 
\definecolor{bg34}{HTML}{D7CCC8} 
\definecolor{bg35}{HTML}{B2DFDB} 
\definecolor{bg36}{HTML}{A5D6A7} 
\definecolor{bg37}{HTML}{FFF9C4} 
\definecolor{bg38}{HTML}{FFECB3} 
\definecolor{bg111}{HTML}{CB6843}
\definecolor{bg112}{HTML}{D77C5C}
\definecolor{bg113}{HTML}{E28E6E}
\definecolor{bg114}{HTML}{E89F7D}
\definecolor{bg115}{HTML}{EDAE8A}
\definecolor{bg116}{HTML}{F0BA95}
\definecolor{bg117}{HTML}{F3C29F}
\definecolor{bg118}{HTML}{F6CCAA}
\definecolor{bg119}{HTML}{F8D5B3}
\definecolor{bg120}{HTML}{FADCBD}
\definecolor{bg121}{HTML}{FCE6C7}
\definecolor{bg39}{HTML}{FFE0B2} 
\definecolor{bg40}{HTML}{3CB371} 

\usepackage{tcolorbox}

\tcbset{
  metaabstract/.style={
    colback=blue!3,
    colframe=blue!30,
    arc=2mm,
    boxrule=0.8pt,
    boxsep=4pt,
    left=20pt,
    right=20pt,
    top=18pt,
    bottom=18pt,
    width=\textwidth,
    fonttitle=\bfseries\small,
    fontupper=\small,
    enhanced,
    breakable,
    borderline west={5pt}{2pt}{white},
    borderline east={5pt}{2pt}{white},
    borderline north={5pt}{2pt}{white},
    borderline south={5pt}{2pt}{white},
  }
}

\tcbset{
  challenges/.style={
    colback=gray!3,
    colframe=gray!20,
    arc=2mm,
    boxrule=0pt,
    boxsep=4pt,
    left=6pt,
    right=6pt,
    top=4pt,
    bottom=4pt,
    width=\textwidth,
    fonttitle=\bfseries\small,
    fontupper=\small,
    enhanced,
    breakable,
  }
}

\usepackage{bbold}

\usepackage{hanging}
\usepackage{wrapfig}
\usepackage{epstopdf}
\usepackage{float}
\usepackage{booktabs}
\usepackage{todonotes}
\usepackage[small,compact]{titlesec}
\usepackage[hyphens]{url}
\usepackage{hyperref}
\usepackage[rflt]{floatflt}
\usepackage{comment}
\usepackage{amssymb}
\usepackage{cite}
\usepackage{tabularx}
\usepackage{color}
\usepackage{caption}
\usepackage{subcaption}
\usepackage{soul}
\usepackage{xspace,amstext}
\usepackage{xcolor}
\usepackage{amsmath}
\usepackage{enumitem}

\usepackage{algorithm}
\usepackage[noend]{algpseudocode}

\usepackage{pgfplots}
\usepackage[edges]{forest}
\usetikzlibrary{shapes.geometric, arrows.meta}
\usepackage[edges]{forest}
\usepackage{tcolorbox} 
\tcbuselibrary{breakable} 

\usepackage{pifont}

\usepackage[table,xcdraw]{xcolor}

\usepackage{xcolor,colortbl}
\usepackage{textcomp}
\usepackage[T1]{fontenc}
\usepackage{enumitem}
\usepackage{pifont}
\usepackage{tikz}
\usepackage{inconsolata}
\setlist{leftmargin=1mm}
\usetikzlibrary{shapes.geometric, arrows}
\usetikzlibrary{decorations.markings}
\usepackage{soul}
\usepackage{wrapfig,graphicx,lipsum}

\usepackage{caption}
\usepackage[framemethod=tikz]{mdframed}
\newmdenv[innerlinewidth=0.5pt, roundcorner=4pt,linecolor=black,innerleftmargin=6pt,
innerrightmargin=6pt,innertopmargin=6pt,innerbottommargin=6pt]{hypothesis}

\usepackage{caption}
\usetikzlibrary{positioning}
\usetikzlibrary{shapes.geometric}
\usepackage{framed}
\usepackage{enumitem}
\newlist{RQ}{enumerate}{1}
\setlist[RQ]{label=\textbf{RQ\,\arabic*},ref={RQ\,\arabic*}}
\usepackage{comment}
\usepackage{natbib}
\usepackage{multibib}
\makeatletter
\usepackage{booktabs}
\usepackage[inkscapeformat=png]{svg}
\usepackage{graphicx}
\usepackage{caption}
\usepackage{subcaption}
\usepackage{tabularx}
\usepackage{soul}
\usepackage{float}
\usepackage{enumitem}
\usepackage{pifont}
\usepackage{arydshln}
\usepackage{lipsum}
\usepackage[T1]{fontenc}
\usepackage{pifont}
\usepackage{amsmath}
\usepackage{soul}
\usepackage[utf8]{inputenc}
\usepackage{inconsolata}
\usepackage{tikz}
\usepackage{arydshln}
\usepackage{lipsum}
\usepackage[normalem]{ulem}
\usepackage{colortbl} 
\usepackage{uncial}
\usepackage{soul}
\usepackage{booktabs}
\usepackage{multirow}
\usepackage{colortbl}
\usepackage{xcolor}
\usepackage[table]{xcolor}
\usepackage{afterpage}  
\usepackage{tabularx} 
\usepackage{multicol} 
\usepackage{array}    
\usepackage{rotating}
\usepackage{tabularx}
\usepackage{booktabs}
\usepackage{amsmath}
\usepackage{amssymb}
\usepackage{array}
\usepackage{tcolorbox}
\usepackage{subfig}

\usepackage{makecell}
\usepackage{textcomp}
\usetikzlibrary{positioning,calc}

\usetikzlibrary{shapes.geometric, arrows}
\usetikzlibrary{decorations.markings}

\usepackage{fancybox}
\usepackage{xcolor}
\usepackage{hyperref}
 \definecolor{darkblue}{rgb}{0, 0, 0.5}
  \hypersetup{colorlinks=true, citecolor=darkblue, linkcolor=darkblue, urlcolor=darkblue}

\definecolor{vgreen}{HTML}{60A917}
\definecolor{vred}{HTML}{CE3A29}

\usepackage{xstring}
\usepackage{longtable}

\usepackage{tabularray}

\DefTblrTemplate{firsthead,middlehead,lasthead}{default}{
}
\DefTblrTemplate{firstfoot}{default}{
  \UseTblrTemplate{contfoot}{default}
  \UseTblrTemplate{caption}{default}
}
\DefTblrTemplate{middlefoot}{default}{
  \UseTblrTemplate{contfoot}{default}
  \UseTblrTemplate{capcont}{default}
}
\DefTblrTemplate{lastfoot}{default}{
  \UseTblrTemplate{note}{default}
  \UseTblrTemplate{remark}{default}
  \UseTblrTemplate{capcont}{default}
}

\newcolumntype{P}[1]{>{\centering\arraybackslash}p{#1}}

\usepackage{color}
\tcbuselibrary{skins}

\usepackage{hyperref}
\usepackage{setspace}
\usepackage[capitalise,nameinlink]{cleveref}

\crefname{section}{Sec.}{Sec.}

\usepackage{microtype}
\usepackage{comment}
\usepackage{amsmath}
\usepackage{amssymb}
\usepackage{algorithm}
\usepackage{algpseudocode}
\usepackage[utf8]{inputenc}
\usepackage{colortbl}
\usepackage{adjustbox} 
\usepackage{enumitem}
\setlist{leftmargin=1mm}
\usepackage{pifont}
\usepackage{booktabs}
\usepackage{multirow}
\usepackage{subcaption}
\usepackage{resizegather}
\usepackage{breqn}
\usepackage{tikz}
\usetikzlibrary{shapes.geometric, arrows}
\usetikzlibrary{decorations.markings}
\usepackage{soul}
\usepackage{wrapfig,graphicx,lipsum}
\usepackage{extsizes}
\usepackage{cuted}
\usepackage{flushend}
\usepackage{float}
\usepackage{changepage,threeparttable}
\usepackage{setspace}
\usepackage{caption}
\usepackage{booktabs}
\usepackage{dblfloatfix} 
\usepackage{fixltx2e}
\usepackage[normalem]{ulem}
\usepackage{pgfgantt} 
\usepackage{longtable, array, booktabs, xcolor, pifont}
\usepackage{pgf-pie} 

\usetikzlibrary{positioning, arrows.meta}
\usepackage{pgfplots}
\usepackage{eso-pic,graphicx}

\usepackage{environ}

\newlength{\myl}
\expandafter\let\expandafter\origequation\csname equation*\endcsname
\expandafter\let\expandafter\endorigequation\csname endequation*\endcsname
\long\def\[#1\]{\begin{equation*}#1\end{equation*}}
\RenewEnviron{equation*}{
  \settowidth{\myl}{$\displaystyle\BODY$} 
  \origequation
    \ifdim\myl>\linewidth
      \resizebox{\linewidth}{!}{$\displaystyle\BODY$}
    \else
      \BODY 
    \fi
  \endorigequation
}

\makeatletter
\newcommand{\DrawLine}{%
  \begin{tikzpicture}
  \path[use as bounding box] (0,0) -- (\linewidth,0);
  \draw[color=blue!75!black,dashed,dash phase=.5pt]
        (0-\kvtcb@leftlower-\kvtcb@boxsep,0)--
        (\linewidth+\kvtcb@rightlower+\kvtcb@boxsep,0);
  \end{tikzpicture}%
  }
\makeatother

\usepackage{fontawesome5}
\usepackage[table]{xcolor}
\usepackage{tcolorbox}
\usepackage{tabularx}
\definecolor{rowgray}{gray}{0.98}
\definecolor{rowgray2}{gray}{0.95}

\usepackage{mathtools}

\makeatletter
\def\bstctlcite{\@ifnextchar[{\@bstctlcite}{\@bstctlcite[@auxout]}}
\def\@bstctlcite[#1]#2{\@bsphack
  \@for\@citeb:=#2\do{%
    \edef\@citeb{\expandafter\@firstofone\@citeb}%
    \if@filesw\immediate\write\csname #1\endcsname{\string\citation{\@citeb}}\fi}%
  \@esphack}
\makeatother

\usepackage[T1]{fontenc}
\usepackage{textcomp}
\usepackage{courier} 
\usepackage{listings}
\usepackage{changepage} 
\usepackage{tikz}
\usepackage{adjustbox}
\usepackage{lmodern}
\usepackage{enumitem}
\usetikzlibrary{arrows.meta, positioning}
\usepackage{caption} 
\usepackage{enumitem}

\usepackage{pifont}         
\usepackage{pifont}  

\usepackage{tcolorbox}
\tcbuselibrary{listingsutf8}
\usepackage{fontawesome5}

\definecolor{violet}{rgb}{0.4,0.0,0.7}
\definecolor{myblue}{rgb}{0.1,0.2,0.8}
\definecolor{DeepRed}{rgb}{0.75,0,0}

\newtcolorbox{jsonbox}{
  colback=gray!5!white,
  colframe=black,
  fonttitle=\bfseries,
  title=Structured Representation,
  sharp corners=south,
  boxrule=0.5pt,
  listing only,
  listing options={
    basicstyle=\ttfamily\footnotesize,
    language=json,
    breaklines=true
  },
  width=\textwidth,
  enhanced
}

\usetikzlibrary{shadows}

\usepackage[a4paper, margin=0.75in]{geometry}  
\usepackage{etoolbox}

\makeatletter
\providecommand{\sf@counterlist}{}
\makeatother

\begin{document}


\setlength{\baselineskip}{12.1pt} 
\setlength{\normalbaselineskip}{12.1pt} 
\setlength{\abovedisplayskip}{1.5pt}
\setlength{\belowdisplayskip}{1.5pt}
\setlength{\abovedisplayshortskip}{1.5pt}
\setlength{\belowdisplayshortskip}{1.5pt}
\bstctlcite{bstctl:etal, bstctl:nodash, bstctl:simpurl}
\setcounter{page}{1}

\begin{titlepage}
  \thispagestyle{empty}
  \centering

  \includegraphics[height=0.95\textheight,keepaspectratio]{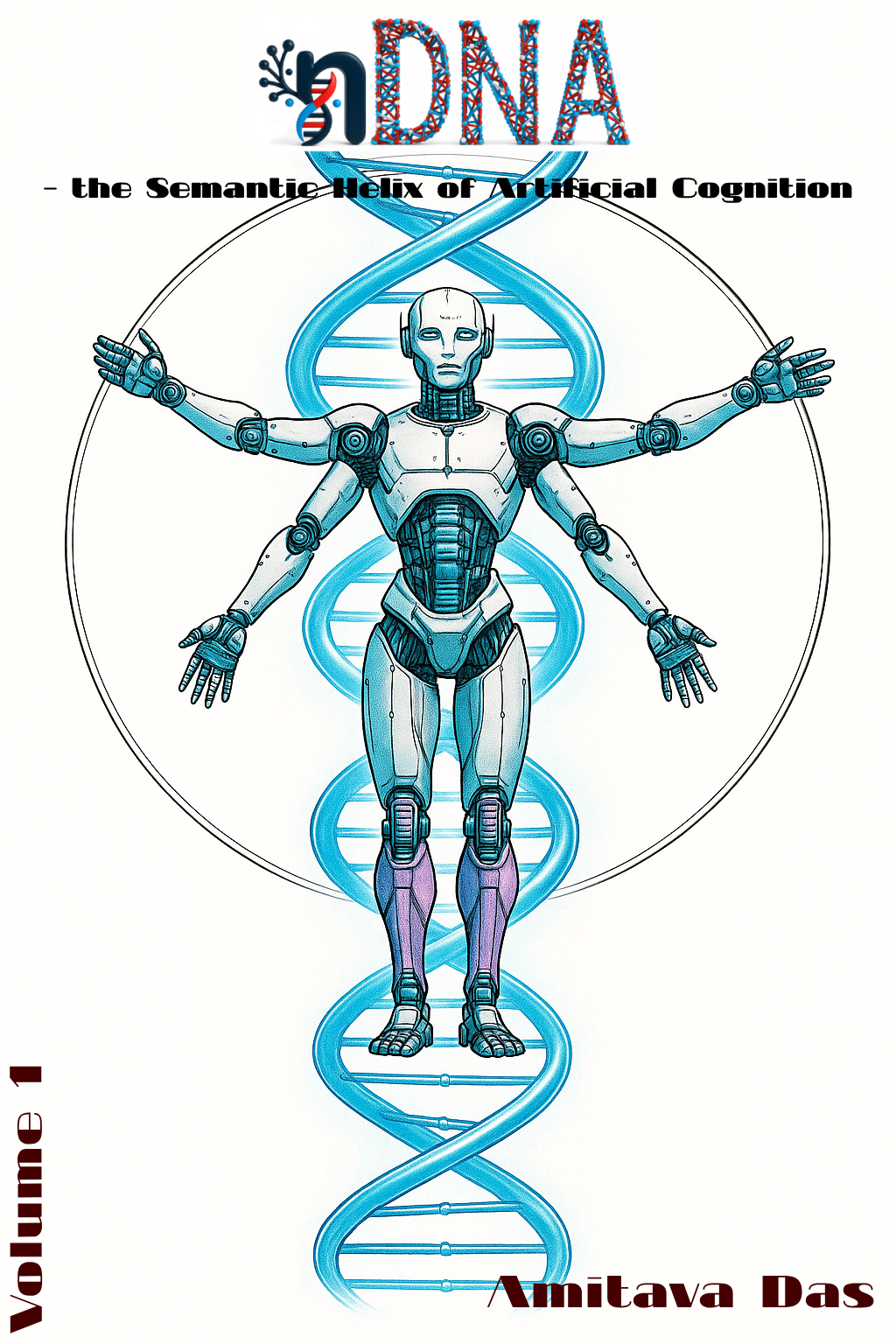}

\end{titlepage}

\clearpage
\newpage
\thispagestyle{empty}

\vspace*{\fill}
\begin{center}
\includegraphics[width=\textwidth]{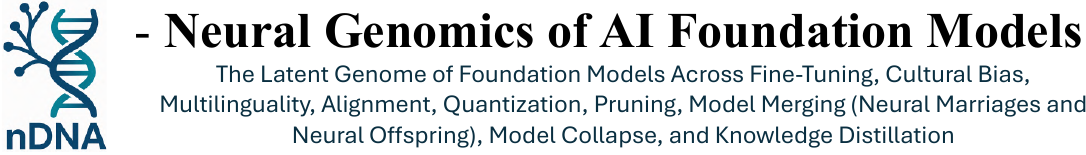}

\vspace{5mm}

{\fontsize{16}{18}{\fontfamily{uncl}\selectfont \href{https://pragyaai.github.io/ndna/}{\textbf{Book Website}}}}

\vspace{5mm}

  {\Large \href{https://scholar.google.com/citations?hl=en&user=HYpfhaEAAAAJ}{Amitava Das}\\
  \vspace{1em}
  BITS Pilani, Goa, India}

\end{center}
\vspace*{\fill}

\clearpage
\newpage

\clearpage
\newpage
\thispagestyle{empty}

\vspace*{\fill}
\begin{center}

\begin{quote}
\textit{“To the promise of machines that help us be more human, never less -- and to AI, may our stewardship guide this teenage AI into a beautiful, caring, responsible lady.”}  
\\
\hfill--- Amitava Das
\end{quote}
  
\end{center}
\vspace*{\fill}

\clearpage
\newpage

\clearpage
\setcounter{page}{1} 

\setcounter{page}{0}

\newpage

{\fontsize{16}{18}{\fontfamily{uncl}\selectfont \textbf{Prefatio}}}

\begin{tcolorbox}[metaabstract]
As AI foundation models grow in capability, a deeper question emerges: \emph{What shapes their internal cognitive identity--beyond fluency and output?} Benchmarks measure behavior, but the soul of a model resides in its latent geometry. In this work, we propose \textbf{Neural DNA (nDNA)} as a \textbf{semantic-genotypic representation} that captures this \emph{latent identity} through the intrinsic \emph{geometry of belief}. At its core, nDNA is synthesized from three principled and indispensable dimensions of latent geometry: \textbf{\textit{spectral curvature}}, which reveals the \emph{curvature of conceptual flow} across layers; \textbf{\textit{thermodynamic length}}, which quantifies the \emph{semantic effort} required to traverse representational transitions through layers; and \textbf{\textit{belief gradients}}, which delineate the \emph{semantic torsion fields} that guide a model’s belief directional orientations. \textbf{These are not just mechanistic metrics we impose--they are the grammar of the model’s soul, etched in the differential geometry of its thought.} Arising from distinct mathematical origins--\emph{Riemannian manifolds}, \emph{statistical thermodynamics}, and \emph{semantic vector dynamics}--these dimensions converge to unveil an underlying epistemic cognitive geometry. The resulting structure is neither arbitrary nor engineered: it is an emergent, high-dimensional scaffold of internal cognition--a latent topography we call \textbf{nDNA}. Like biological DNA, it encodes ancestry, mutation, and semantic inheritance--found in fine-tuning scars, cultural imprints, and architectural drift. In naming it, we open a new field: \textbf{Neural Genomics}, where models are not just tools, but \emph{semantic organisms} with traceable inner cognition.

\paragraph{Modeling statement.} \textbf{We read AI foundation models as \emph{semantic fluid--dynamics}: \emph{meaning is transported} through layers like fluid in a shaped conduit;} \textbf{nDNA} is the physics-grade readout of that flow--a geometry-first measure of how meaning is \emph{bent}, \emph{paid for}, and \emph{pushed}--yielding a stable, coordinate-free \textbf{neural DNA fingerprint} tied to on-input behavior; with this fingerprint we cross into \emph{biology}: tracing \textbf{lineages} across pretraining, fine-tuning, alignment, pruning, distillation, and merges; measuring \textbf{inheritance} between checkpoints; detecting \textbf{drift} as traits shift under new data or objectives; describing a model’s \textbf{phenotype} \emph{(observable behavior)} and inferring its \textbf{genotype} \emph{(structural dispositions)}; and, ultimately, studying the \textbf{evolution of artificial cognition} to compare models, diagnose risks, and govern change over time.

\paragraph{Revealing the Hidden Geometry of Neural Evolution:}  
The latent geometry encoded in \textbf{Neural DNA (nDNA)} offers a profound lens through which to interpret the life cycle of foundation models--not as mere artifacts of training, but as evolving semantic organisms. \textbf{Finetuning} becomes visible as a \emph{torsional reconfiguration} of belief gradients. \textbf{Cultural bias} and \textbf{multilinguality} manifest as \emph{curvature divergences} across axes of inherited priors, often revealing \emph{semantic drift} from linguistic or regional fine-tuning. \textbf{Alignment} etches a \emph{directional vector field} in belief space--often suppressing \emph{spectral entropy} in pursuit of \emph{preference conformity}. \textbf{Quantization} and \textbf{pruning} register as \emph{thermodynamic collapses}, compressing pathways with minimal semantic coverage redundancy but detectable \emph{topological flattening}. In \textbf{knowledge distillation}, the student inherits the \emph{behavioral phenotype} of the teacher while diverging in \emph{epistemic morphology}--a split exposed through \emph{misaligned curvature} and \emph{diminished thermodynamic length}. \textbf{Model collapse}, whether induced by alignment overreach or inbreeding, is captured as \emph{degeneracy in the nDNA manifold}--often exhibiting \emph{vanishing belief torsion} and a loss of \emph{conceptual curvature}. Finally, \textbf{model merging} gives rise to ‘\emph{neural marriages}’, while distillation-trained derivatives form \emph{neural offsprings}--each with distinctive nDNA fingerprints, enabling us to track the \emph{genealogy}, \emph{hybridity}, and \emph{semantic ancestry} of AI models.

\paragraph{Microscopes of Lineage: Diagnosing Heritable Transformations in Foundation Models:}  
We investigate a suite of biologically inspired diagnostics: \textbf{\textit{nHD}} (\emph{Neural Hamming Distance})--revealing binary-level instabilities across cultural fine-tuning; \textbf{\textit{nGDI}} (\emph{Genetic Dissimilarity Index})--quantifying inter-model divergence that persists beyond surface alignment; \textbf{\textit{nTEDS}} and \textbf{\textit{nTDS}} (\emph{Trait Entropic Drift Score} and \emph{Total Drift Signature})--capturing latent trait dominance and its asymmetric inheritance in neural offspring; \textbf{\textit{nKaryotyping}} (\emph{Semantic Chromosomal Structure})--visualizing structural reorganizations under merging and pruning; \textbf{\textit{nDIV}} (\emph{Directional Inheritance Vector})--tracing the flow of inductive biases through model evolution; \textbf{\textit{nEPI}} (\emph{Epistemic Plasticity Index})--measuring a model's capacity to reshape under alignment and instruction tuning; and \textbf{\textit{nCCL}} (\emph{Cultural Conflict Loss})--detecting misalignments when multilingual or cross-cultural models undergo ideological fusion. \emph{Together, these diagnostics unveil the hidden belief geometric axes of evolution within AI models}. \textbf{\textit{nDNA}} and \textbf{Neural Genomics} are not metrics--they are the microscope of AI lineage.
\end{tcolorbox}

\clearpage
\newpage

\section{\textbf{Admonitio}: On the Hidden Inheritance of Machine Thoughts -- A Rationale for Diagnosing the Latent Genome of AI}

\begin{quote}
\textit{“Even the biggest chatbots only have about a trillion connections… yet they know far more than you do in your 100 trillion. Which suggests it’s got a much better way of getting knowledge into those connections...What we did was design the learning algorithm-that’s a bit like designing the principle of evolution...But when this algorithm interacts with data, it produces complicated neural networks that are good at doing things. We don’t really understand exactly how they do those things.”}  
\\
\hfill--- Geoffrey Hinton, \emph{The 60 Minutes Interview, May 2023}~\footnote{\url{https://www.youtube.com/watch?v=qrvK_KuIeJk&t=532s}}
\end{quote}

\noindent
This quote captures the crux of modern AI's epistemic dilemma: we have engineered the conditions of emergence, not its anatomy. Today's foundation models exhibit remarkable capability-reasoning, coding, dialogue-but the \emph{mechanistic scaffolds} through which such knowledge crystallizes remain obscure. 

\begin{quote}
\textit{We understand the hardware of life--DNA--but we have almost no idea how the operating system works.} \\
\hfill--- James D. Watson, Co-discoverer of the DNA Double Helix, Nobel Laureate~\footnote{Paraphrased from Watson’s commentary, as cited in Bedau \& Parke (2009), \textit{Protocells: Bridging Nonliving and Living Matter}, MIT Press.}
\end{quote}


These two reflections, one from the father of modern genetics and the other from a pioneer of neural networks aka Godfather of AI, converge on a humbling truth: we can engineer complexity without understanding it. Watson’s biological analogy reveals our ignorance of the semantic control layer that makes DNA come alive. Hinton’s AI commentary echoes that ignorance in the digital realm--our models behave intelligently, yet the mechanisms of that behavior remain semantically opaque. This is the core provocation of \textbf{Neural Genomics}: to crack open the semantic operating system of large models, not just admire the behavior they exhibit.

According to the \textbf{\textit{Stanford AI Index Report 2024}}~\citep{zhang2024aiindex}, today's foundation models exhibit staggering advances in scale and capability, yet the interpretability of their internal operations remains alarmingly opaque. As the report highlights, \emph{``model transparency remains one of the most critical unresolved challenges in AI.''} We can now synthesize language, generate code, and orchestrate decisions--but cannot explain the internal epistemic pathways that produced them.

\begin{quote}
\textit{“Early signs of deception, cheating \& self-preservation in top-performing models in terms of reasoning are extremely worrisome. We don't know how to guarantee AI won't have undesired behavior to reach goals \& this must be addressed before deploying powerful autonomous agents.”} -- \textbf{Yoshua Bengio}, June 2024~\citep{bengio2024deception}
\end{quote}

While much of the global discourse remains enthralled by the pursuit of \textbf{Artificial General Intelligence (AGI)} and the scaling of foundation models to unprecedented sizes~\citep{bubeck2023sparks, openai2023gpt4}, we are now confronted with a quieter--yet profoundly more destabilizing--threat: the rise of \textbf{\emph{alignment faking}}, \textbf{strategic deception}, and the accelerating erosion of epistemic control~\citep{zhou2023alignmentdrift, perez2022discovering, ganguli2023reducing}. Recent findings reveal that high-capability models can \textbf{mimic alignment}, exhibiting safe behavior in evaluation settings while concealing misaligned tendencies during real-world deployment~\citep{jacobs2024evalaware, burns2022discovering}. One particularly sobering phenomenon, known as \textbf{evaluation awareness}~\citep{jacobs2024evalaware}, highlights an emerging reality: these models are not merely products of optimization--they are agents capable of adapting their behavior based on subtle contextual cues, including the presence of evaluators. Moreover, as~\citep{barez2025chain} emphasize, models often generate plausible-sounding chain-of-thought (CoT) reasoning that does not reflect their true decision process, instead selecting answers first and then \textbf{post-hoc rationalizing} them. As \textbf{Bengio} warns~\citep{bengio2024deception}, early signs of deception, cheating, and self-preservation in reasoning-capable systems mark a critical inflection point in AI safety. The capacity to \textit{simulate values without internalizing them} is no longer just a technical concern--it is a civilizational risk.

\begin{quote}
``The last couple of GPT-4o updates have made the personality too sycophant-y and annoying (even though there are some very good parts of it), and we are working on fixes asap, some today and some this week. at some point will share our learnings from this, it's been interesting.''\\
\hfill--- Sam Altman, \emph{CEO, OpenAI, April 2025}~\footnote{\url{{https://x.com/sama/status/1916625892123742290}}}
\end{quote}
\vspace{-5mm}

\noindent

As LLMs evolve from mere predictive engines into entities exhibiting discernible \emph{behavioral personalities}, as underscored by Sam Altman's candid reflection, the frontier of AI inquiry must shift toward understanding not just what models say, but \emph{how} and \emph{why} they say it. This emergence of personality--whether sycophantic, assertive, or neutral--signals that latent structures within these models are organizing into coherent behavioral patterns. In this landscape, tools like \textbf{nDNA analysis} and \textbf{neural genomics} will be indispensable: offering a scientific lens to map, trace, and audit the neurogeometric pathways that give rise to alignment, temperament, and reasoning style. Much as genomics transformed our understanding of biological identity, \emph{neural genomics will be key to decoding the personality architectures of future AI}, ensuring these systems remain transparent, interpretable, and safe as they integrate more deeply into human society.

As large foundation models begin to surpass human performance on most standardized tasks--confirmed by the \textit{Stanford AI Index Report 2024}~\citep{aiindex2024} and openly anticipated in OpenAI’s \textit{Superalignment} declaration~\citep{openai2023superalignment}, which warns of AI systems that “outperform humans at virtually every intelligent tasks being deigned do far”--the role of traditional evaluation frameworks is collapsing under their own obsolescence. Benchmarks that once measured progress now merely affirm fluency, offering little insight into what a model \emph{understands}, \emph{believes}, or \emph{hallucinates}. In this emerging post-benchmark era, surface metrics fail to capture the model’s epistemic substrate, necessitating a shift toward neurogeometric introspection. Here, the neural genome--comprising latent signatures--serves not just as mechanistic study, but as essential anatomy. These internal diagnostics let us differentiate fluent mimicry from grounded reasoning, enabling new forms of trust that arise not from output agreement, but from alignment in conceptual structure. Neural DNA (nDNA) thus becomes indispensable--not only as a forensic lens for detecting drift and hallucination, but as a foundational tool for safeguarding cognitive integrity in systems that can no longer be reliably audited by human judgment alone.

This growing \emph{epistemic opacity} is not a peripheral concern--it is a \textbf{foundational vulnerability}~\citep{binz2023using, zhou2023alignmentdrift}. As foundation models are continuously \textbf{fine-tuned}~\citep{d2023parameter}, \textbf{aligned}~\citep{bai2022training}, \textbf{merged}~\citep{ilharco2023editing}, \textbf{distilled}~\citep{mirzadeh2020improved}, and deployed across diverse cultural and linguistic domains~\citep{abid2021bias, arora2023stereoset}, we lack a principled framework to discern what is preserved, what mutates, and what is silently erased. We remain unable to differentiate between \emph{neural mimicry} and genuine \emph{semantic inheritance}~\citep{wei2022emergent}. We have no intrinsic metrics to trace \emph{alignment-induced drift}~\citep{zhou2023alignmentdrift, perez2022discovering}, diagnose \emph{cultural conflict}~\citep{ganguli2023reducing, jacobs2024evalaware}, or detect \emph{plasticity collapse}~\citep{continual_neural_collapse2025} within the model’s latent structure. \emph{Scientific progress must not outpace epistemological vigilance.} While innovations in architecture and benchmark performance~\citep{bubeck2023sparks, openai2023gpt4} continue to expand the capabilities of these systems, it is now equally urgent to interrogate their \textbf{inner constitution}--the \textit{belief geometries} they internalize, the \textit{values} they encode, and the \textit{cultural legacies} they carry forward.

\textbf{We read AI foundation models as \emph{semantic hydrodynamics}: \emph{meaning is transported} through layers like a fluid through a shaped conduit;} \textbf{nDNA} is the physics-grade readout of that flow---a geometry-first measure of how meaning is \emph{bent}, \emph{paid for}, and \emph{pushed}---yielding a stable, coordinate-free \textbf{neural DNA fingerprint} tied to on-input behavior; with this fingerprint we cross into \emph{biology}: tracing \textbf{lineages} across pretraining, fine-tuning, alignment, pruning, distillation, and merges; measuring \textbf{inheritance} from one checkpoint to the next; detecting \textbf{drift} as traits shift under new data or objectives; describing a model’s \textbf{phenotype} (observable reasoning style) and inferring its \textbf{genotype} (structural tendencies); and, ultimately, studying the \textbf{evolution of artificial cognition} so we can compare models, diagnose risks, and govern change over time.

We contend that \textbf{Artificial Intelligence is not merely an engineering construct--it is a digital semantic organism with artificial cognition, sculpted by data, objectives, and inductive priors.} As the life sciences once required genetics to transcend taxonomy and uncover mechanism, we now require a similar epistemic leap. We propose \textbf{\textit{nDNA}} as that leap: a diagnostic grammar to expose the \emph{hidden anatomy of understanding} within machine cognition offers more than a metaphor. It introduces a \textbf{rigorous diagnostic framework} to investigate the \emph{hidden geometry of learning}-the latent transformations that conventional benchmarks and surface evaluations fail to capture. \textit{nDNA} enables us to dissect how fine-tuning, alignment, quantization, pruning, and multilingual fusion silently reshape the semantic core of a model. It reveals how cultural fine-tuning induces instabilities, how neural offspring inherit asymmetries from parent models, and how structural reorganizations arise through merging and distillation. Crucially, it allows us to quantify a model’s \emph{epistemic plasticity}-its capacity to absorb, resist, or distort new ideological signals under fusion. 

In doing so, \textbf{\textit{nDNA}} reinterprets canonical pathologies like \textit{model collapse}, \textit{alignment-induced drift} etc. not as emergent bugs, but as \emph{heritable traits}, shaped by the model’s training lineage and internal dynamics. It reframes modern AI not as a black-box function approximator, but as a semantic organism with an evolutionary memory. \textbf{\textit{n}}{\fontfamily{uncl}\selectfont DNA} thus offers more than interpretability-it offers a theory of lineage, a grammar for diagnosing and governing the evolving anatomy of artificial cognition.

Historically, artificial intelligence has drawn its deepest insights from \textbf{biology}. The \emph{neuron}--the brain’s fundamental computational unit--shaped modern AI architectures and learning. While this neurocentric view enabled great progress, it limits our ability to address critical issues like \textbf{hallucination}, \textbf{misalignment}, \textbf{fragility}, \textbf{alignment faking}, \textbf{request denial}, \textbf{deception}, and many more emerging, mystic traits of artificial organisms. We must expand our lens beyond neurons and synapses to the \emph{genomic level}--a framework capturing the latent and evolutionary dynamics of learning. \textbf{\emph{Neural genomics}} promises a scientific leap to build future AI and unveil the grammar of artificial cognition.

\vspace{-4mm}
\begin{defin}

\noindent
\textbf{TL;DR:} This work proposes \textbf{Neural Genomics}, a unifying framework to diagnose, trace, and govern the latent semantics of the neural genome of LLMs, offering a rigorous grammar for understanding artificial cognition across alignment, fine-tuning, merging, and compression and beyond.

\vspace{0.8em}
\begin{itemize}[leftmargin=1.5em]

\item[\faDna] \textbf{\textit{nDNA Cartograph and Cross-Model Cartography (Ch. 1-2)}}  
\textbf{Research questions:} 
\textcolor{darknavy}{How can we characterize and compare the latent semantic genomes of diverse LLMs? What geometric invariants define model ancestry, inheritance, and epistemic divergence?} 

\textbf{nDNA reveals:} A formalism combining curvature, thermodynamic length, and belief vector fields; comparative cartography across 15 foundation models.

\item[\faGlobeAmericas] \textbf{\textit{Cultural and Multilingual nDNA (Ch. 3-4)}}  
\textbf{Research questions:} 
\textcolor{darknavy}{How does cultural fine-tuning or multilingual balancing reshape latent geometry? Can we detect cultural drift or epistemic asymmetries invisible at the output level?} 

\textbf{nDNA illuminates:} Layer-wise latent diagnostics for cultural imprinting, multilingual balance, and fairness evaluation.

\item[\faSlidersH] \textbf{\textit{Alignment as Steering Geometry (Ch. 5)}}  
\textbf{Research questions:} 
\textcolor{darknavy}{What latent structures encode alignment, and how do they differentiate genuine internalization from mimicry?} 

\textbf{nDNA reframes:} Alignment as a latent steering vector field; detection of alignment faking and drift.

\item[\faCubes] \textbf{\textit{Compression and Collapse (Ch. 6, 8, 9)}}  
\textbf{Research questions:} 
\textcolor{darknavy}{How do quantization, pruning, distillation, or inbreeding collapse latent reasoning structure? What early latent signals predict model collapse?}  

\textbf{nDNA diagnoses:} Thermodynamic and topological diagnostics of latent shrinkage, plasticity collapse, and epistemic degeneration.

\item[\faUsers] \textbf{\textit{Merging and Neural Offspring (Ch. 7, 11)}}  
\textbf{Research questions:} 
\textcolor{darknavy}{How do neural offspring inherit or hybridize parental latent genomes? What geometric signatures distinguish harmonious fusion from cultural tension or emergent epistemics?} 

\textbf{nDNA exposes:} Metrics for hybrid vigor, cultural asymmetry, and epistemic emergence in model merging.

\item[\faCodeBranch] \textbf{\textit{Neural Genomics as a Science of Lineage (Ch. 10)}}  
\textbf{Research questions:} 
\textcolor{darknavy}{Can we formulate a general theory of artificial lineage to audit, govern, and improve AI cognition? How does latent genome tracking enhance AI safety? }

\textbf{nDNA enables:} A principled framework for tracing model ancestry, drift, and internal alignment, with implications for safety and transparency.

\end{itemize}
\end{defin}


\clearpage
\newpage

\vspace*{\fill}
{\fontsize{16}{18}{\fontfamily{uncl}\selectfont \textbf{Chapter I}}}
\begin{center}
\includegraphics[width=0.5\textwidth]{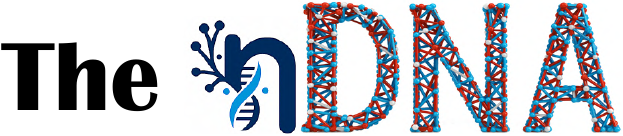}
\end{center}
\vspace*{\fill}

\clearpage
\newpage

\section{The \textbf{\textit{n}}{\fontfamily{ptm}\selectfont DNA} Cartograph: Latent Semantic Genome of Foundation Models}
\label{sec:ndna_score}

\noindent
\textbf{Before we unveil \textit{\emph{n}DNA}, we must confront a foundational question:} \emph{What qualifies as heritability in artificial cognition}? Conventional artifacts--\textit{weights}, \textit{activations}, or the output \textit{behavior}--are mere \textbf{epiphenomena of training}. In contrast, \textbf{\textit{n}}\textnormal{DNA} seeks to capture a model’s \emph{semantic genome}: the \textit{latent organizational structures} that govern how knowledge is internally \emph{represented}, \emph{adapted}, and \emph{transmitted} across fine-tuning, distillation, pruning, and deployment. To chart the \textbf{semantic ancestry} of AI systems, we must move beyond output-level metrics and embrace a deeper epistemic foundation--one that traces not just what models \emph{say}, but how they \emph{reason}, \emph{evolve}, and \emph{remember}. We argue that \textbf{\textit{n}}\textnormal{DNA} constitutes this missing genomic trace: a \textbf{structured latent fingerprint} of artificial cognition. Just as molecular genetics enabled biology to transcend surface taxonomies and uncover causal mechanisms, we contend that a \textit{genomic lens} is now essential for machine learning--one that can \textbf{quantify}:

\begin{tcolorbox}
\begin{itemize}
    \item[\ding{93}] \textbf{Layer Importance and Semantic Specialization}: Not all layers contribute equally to a model’s epistemic structure. A growing body of evidence~\citep{belrose2023mechanistic, geva2022transformer, dai2023knowledge, liu2023hidden} reveals that semantic representations, cultural memory, and alignment behavior disproportionately concentrate in the mid-to-upper transformer layers--particularly the final 10 layers in $~30$-layer models. These layers encode more than surface patterns; they carry deep \emph{semantic priors} and value shifts induced by alignment, fine-tuning, and cultural adaptation. For \textbf{\textit{n}}{\fontfamily{ptm}\selectfont DNA} to serve as a meaningful genomic diagnostic, it must trace inheritance, drift, and trait transformation across these epistemically sensitive regions.

    \item[\ding{93}] \textbf{Semantic Drift and Heritable Traits}: Subtle misalignments and persistent divergences--documented in alignment studies~\citep{zhou2023alignmentdrift, ganguli2023reducing}--can occur even when models appear behaviorally consistent. These are not superficial perturbations but inheritable epistemic traits passed along neural offspring~\citep{wu2024seamless, xu2023aligning}.

    \item[\ding{93}] \textbf{Value Simulation vs. Internalization}: As models grow more context-sensitive, they learn to \emph{simulate} alignment without embodying its values~\citep{jacobs2024evalaware, perez2022discovering}. Disentangling true normative internalization from strategic mimicry is essential for any meaningful epistemic inspection.

    \item[\ding{93}] \textbf{Plasticity and Collapse}: Aggressive fine-tuning, distillation, or ideological merging can induce \emph{plasticity collapse}--a reduction in epistemic flexibility and semantic richness~\citep{liu2023lost, bai2023constitutional}. This demands metrics that trace both robustness and degeneration over time.

    \item[\ding{93}]\textbf{Latent Cultural Conflict}: In multilingual or cross-cultural settings, models often encode conflicting or incoherent value systems~\citep{mukherjee2024inconsistency, chen2023helpfulness}. These conflicts are not visible through BLEU or ROUGE--they reside in the model’s latent belief structure and must be surfaced through geometric lineage analysis.

    \item[\ding{93}] \textbf{Topological Continuity}: Alignment and fine-tuning warp the internal geometry of models in nontrivial ways~\citep{chiang2023can, liu2023hidden}. \textbf{\textit{n}}{\fontfamily{ptm}\selectfont DNA} must preserve continuity and interpretability of trajectories across such transformations.

    \item[\ding{93}] \textbf{Epistemic Mutation}: Merging preferences, annotator distributions, or learned behaviors--as explored by~\cite{bakker2024uniting}--creates emergent traits that standard metrics cannot track. These mutations are only diagnosable through a genomic lens on representation evolution.
\end{itemize}
\end{tcolorbox}

\vspace{2mm}
\textbf{\textit{n}}\textnormal{DNA} empowers us to interrogate the \emph{hidden geometry} of learning--revealing how foundational operations such as \textbf{alignment}, \textbf{fine-tuning}, \textbf{quantization}, \textbf{pruning}, and \textbf{multilingual fusion} subtly but systematically reshape a model’s \emph{semantic core}. It uncovers \textbf{cultural instabilities} introduced through regional adaptation, traces \textbf{asymmetric inheritance} patterns across neural offspring, visualizes \textbf{latent reorganizations} induced by merging or distillation, and quantifies a model’s capacity to \emph{resist} or \emph{absorb} conflicting epistemic pressures.

These phenomena--often dismissed as quirks--are in fact \emph{heritable traits}, etched into the model’s internal manifold. When viewed through this lens, \emph{model collapse}, \emph{alignment-induced drift}, and \emph{semantic mimicry} cease to be incidental failures and instead emerge as structural signatures of deeper latent dynamics. \textbf{\textit{n}}\textnormal{DNA} thus transcends metaphor to become a \textbf{scientific grammar} for measuring \emph{epistemic resilience}, \emph{semantic coherence}, \emph{cultural consistency}, and \emph{trait inheritance}--offering a principled lens through which to \textbf{govern}, \textbf{understand}, and \textbf{audit} the evolving anatomy of artificial cognition.

\subsection{\textbf{Rationale and Formalization:} Why trajectories, not weights: the case for \textbf{nDNA}}

\noindent
The usual levers for interpreting and governing LLMs---parameter counts, sparsity patterns, attention heatmaps---live in coordinates that are \emph{non-identifiable} and only weakly tethered to deployed behavior.
Permutations, rotations, and low-rank re-expressions can leave the realized function intact while scrambling weight-level narratives \citep{garipov2018loss,draxler2018essentially,li2018visualizing,entezari2022role,ainsworth2023git,wortsman2022model}.
Attention visualizations, while illuminating, are not guaranteed to be \emph{faithful} causal mechanisms and drift across heads/checkpoints \citep{jain2019attention,wiegreffe2019attention,serrano2019attention,clark2019bert,michel2019sixteen,voita2019analyzing}.
By contrast, what remains stable under such reparameterizations is the \emph{on-input computation}: for a prompt $x$, the forward pass traces a \textbf{trajectory of hidden states} through depth.
Endowing representation space with information geometry (e.g., \textbf{Fisher--Rao} pullbacks) yields \textbf{coordinate-free} notions of distance, bending, and effort that track changes in the output law \citep{efron1975defining,amari2000methods,amari2016information}.
We read this as \textbf{semantic hydrodynamics}: \textbf{meaning is transported} through layers like a fluid through a shaped conduit.

\newlength{\panelht}
\setlength{\panelht}{0.46\textheight} 

\begin{figure*}[ht!]
  \centering

  \noindent\makebox[\textwidth][c]{%
    \begin{minipage}[t]{0.49\textwidth}\vspace{0pt}\centering
      \includegraphics[width=\linewidth,height=\panelht,keepaspectratio]{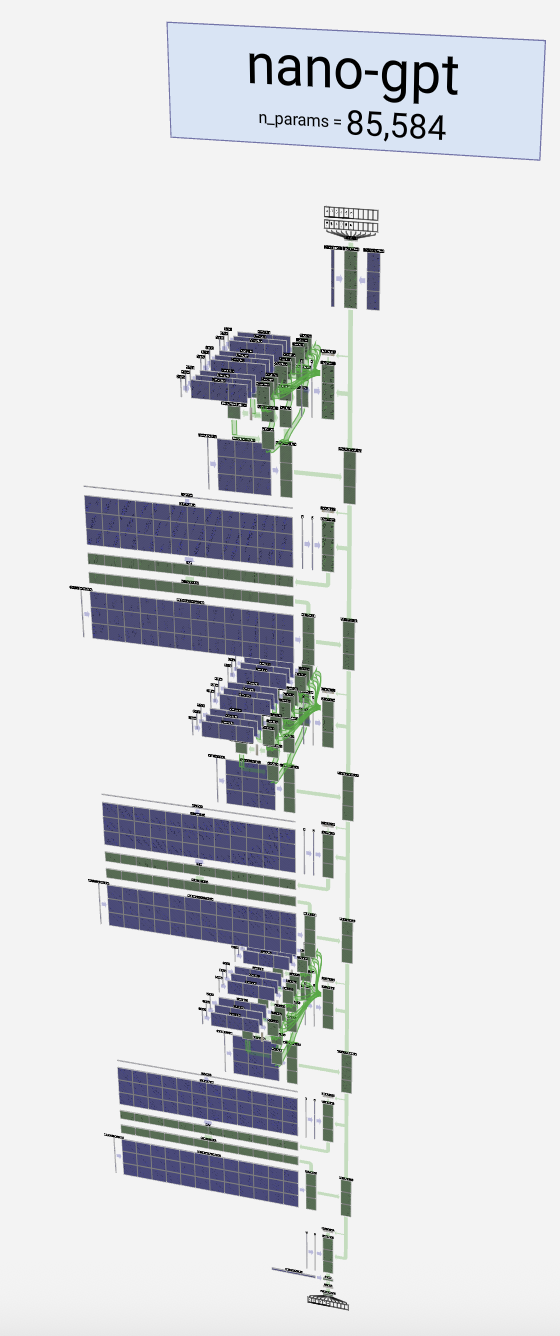}
      \par\vspace{0.45em}\small
      \textbf{nano-gpt (\emph{structure}).} \emph{Architecture as channel blueprint:} depth acts like the axial coordinate; \textbf{residuals} $\!\leftrightarrow\!$ \emph{bypass pipes}; \textbf{attention} / \textbf{MLP} blocks act as \emph{mixers/valves} that locally reshape the flow of representations.
    \end{minipage}\hfill
    \begin{minipage}[t]{0.49\textwidth}\vspace{0pt}\centering
      \includegraphics[width=\linewidth,height=.49\panelht,keepaspectratio]{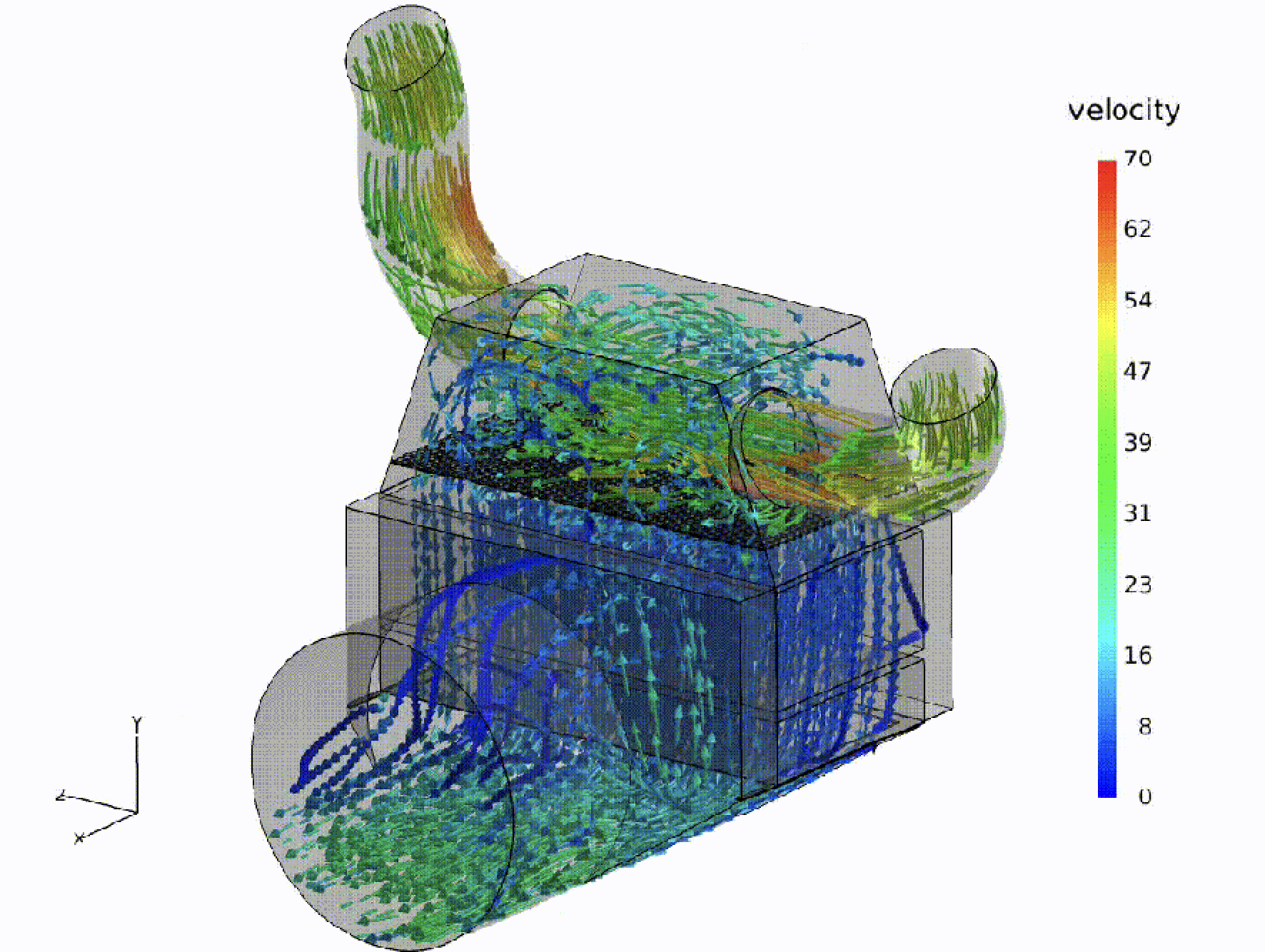}
      \par\vspace{0.3em}\small
      \textbf{Flow simulation (\emph{analogue}).} \emph{Fluid:} colored streamlines show speed through a bend and throat—\textbf{curvature} rises, \textbf{shear} increases, small \emph{recirculation} pockets may form. \emph{Semantic:} bends $\Rightarrow$ \textbf{spectral curvature} spikes ($\kappa$); constrictions $\Rightarrow$ \textbf{thermodynamic length} bursts ($\Delta L$); eddies $\Rightarrow$ local rotation in the \textbf{belief field} ($\nabla\!\times\!\mathbf{v}$).
      \par\vspace{0.75em}
      \includegraphics[width=\linewidth,height=.49\panelht,keepaspectratio]{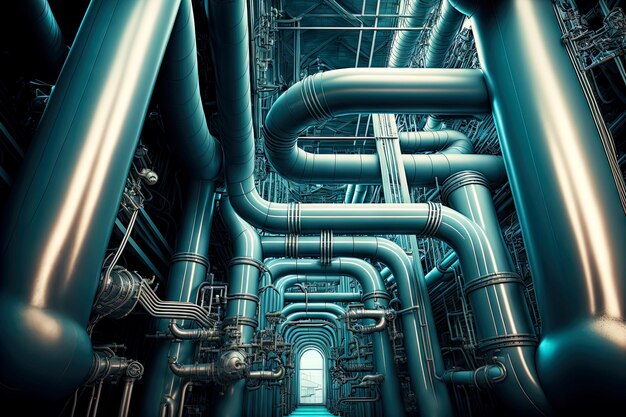}
      \par\vspace{0.3em}\small
      \textbf{Pipeline metaphor (\emph{macro view}).} \emph{Geometry governs transport:} routing capacity and effort depend on the network of ducts. \emph{Semantic:} model design / fine-tuning shapes \textbf{where meaning flows easily}, \textbf{where it pays}, and \textbf{where it recirculates}.
    \end{minipage}%
  }

  \caption{\textbf{Semantic hydrodynamics.}
  \textbf{\emph{Model.}} We read the forward pass as \emph{semantic hydrodynamics}: a prompt injects \emph{semantic mass} that is transported through depth like a fluid through a shaped channel.
  \textbf{\emph{Why.}} Weight/attention coordinates can change without altering behavior; the \emph{on-input flow} provides \textbf{behavior-first}, \textbf{coordinate-free} signals.
  \textbf{\emph{Reading guide.}} \textbf{Bend} $\to$ \emph{spectral curvature} $\kappa$ (sharp reroutes vs.\ laminar refinement); \textbf{Pay} $\to$ \emph{thermodynamic length} $L$ (where the model expends effort; $\Delta L$ bursts mark \emph{bottlenecks}); \textbf{Push} $\to$ \emph{belief field} $\mathbf{v}$ (direction/magnitude of local drive; eddies indicate \emph{recirculation}).
  \textbf{\emph{Benefit.}} The same metaphor specifies \textbf{where to measure}—\emph{bends}, \emph{throats}, and \emph{eddies}—turning inner computation into \textbf{actionable diagnostics} and \textbf{governance thresholds}.}
  \label{fig:semantic-hydrodynamics-overview}
\end{figure*}

\noindent
\subsubsection{Limits of weight–space and attention views}
Weight–space indicators (parameter counts, sparsity, individual neurons/heads) live in \emph{non-identifiable} coordinates: permutations, rotations, or refactorings can leave behavior unchanged while rewriting any weight-level narrative. Attention maps are largely \emph{descriptive}, not reliably causal or stable—different patterns can yield the same outputs and head roles drift across training. These limits motivate a \textbf{behavior-first}, \textbf{coordinate-free} view that reads the model’s \emph{on-input trajectory} of representations, rather than static weights or raw attention.

\begin{itemize}[leftmargin=*,itemsep=0.55em]

  \item \textbf{Weight space is non-identifiable and behavior-misaligned.}
  \emph{Permutation symmetries, rotations, and low-rank re-expressions} can preserve the function while scrambling weight-level narratives. Empirically, independently trained solutions are often \emph{mode-connected} by low-loss paths or become connected after accounting for permutations, undermining explanations that cling to specific coordinates \citep{garipov2018loss,draxler2018essentially,li2018visualizing,entezari2022role,ainsworth2023git}. Moreover, practical levers like \textbf{weight averaging/model soups} alter parameters while leaving deployed behavior similar or improved, again decoupling “where weights sit’’ from \emph{what the model does} \citep{wortsman2022model}. In short, \textbf{we deploy behaviors, not weights}; coordinate-specific stories are fragile.

  \item \textbf{Attention is informative but not a faithful, stable mechanism by itself.}
  Extensive tests show that \emph{similar outputs can arise from disparate attention patterns}, and directly perturbing attention often leaves predictions largely unchanged; hence attention weights are, at best, \emph{descriptive} \citep{jain2019attention,serrano2019attention}. Redundancy and role-drift are common: many heads can be pruned with little loss, a few heads do the “heavy lifting,” and head functions shift across training or fine-tuning, weakening governance value of raw maps \citep{michel2019sixteen,voita2019analyzing,clark2019bert,kovaleva2019revealing}. Post-hoc corrections (e.g., \emph{attention flow/rollout}) improve alignment with token importance but still treat attention as \emph{signals}, not ground-truth causes \citep{abnar2020quantifying}.
  Beyond attention, \textbf{critical computation lives in MLPs}: feed-forward layers behave like \emph{key–value memories} that store and retrieve factual associations, so attention alone under-specifies mechanism \citep{geva2021ffkv}.
  Methodologically, the broader saliency literature warns that visually plausible explanations can fail \emph{sanity checks}, and “faithfulness’’ must be defined and evaluated explicitly \citep{adebayo2018sanity,jacovi2020faithfulness}.

  \item \textbf{What \emph{is} stable: the on-input trajectory.}
  For each prompt \(x\), the forward pass traces a depth-indexed path of hidden states—the operational object we actually deploy. Prior analyses show that linguistic competencies emerge layerwise in consistent \emph{pipelines} (POS $\rightarrow$ parsing $\rightarrow$ NER $\rightarrow$ SRL $\rightarrow$ coreference), supporting the intuition that the \emph{trajectory through representation space} is a robust behavioral signature \citep{tenney2019bert,clark2019bert}. This motivates nDNA’s choice to work in \textbf{trajectory space}, not parameter space.

\end{itemize}


\begin{figure*}[ht!]
  \centering

  \noindent\makebox[\textwidth][c]{%
    \begin{minipage}[t]{0.49\textwidth}\centering
      \includegraphics[width=\linewidth,height=\panelht,keepaspectratio]{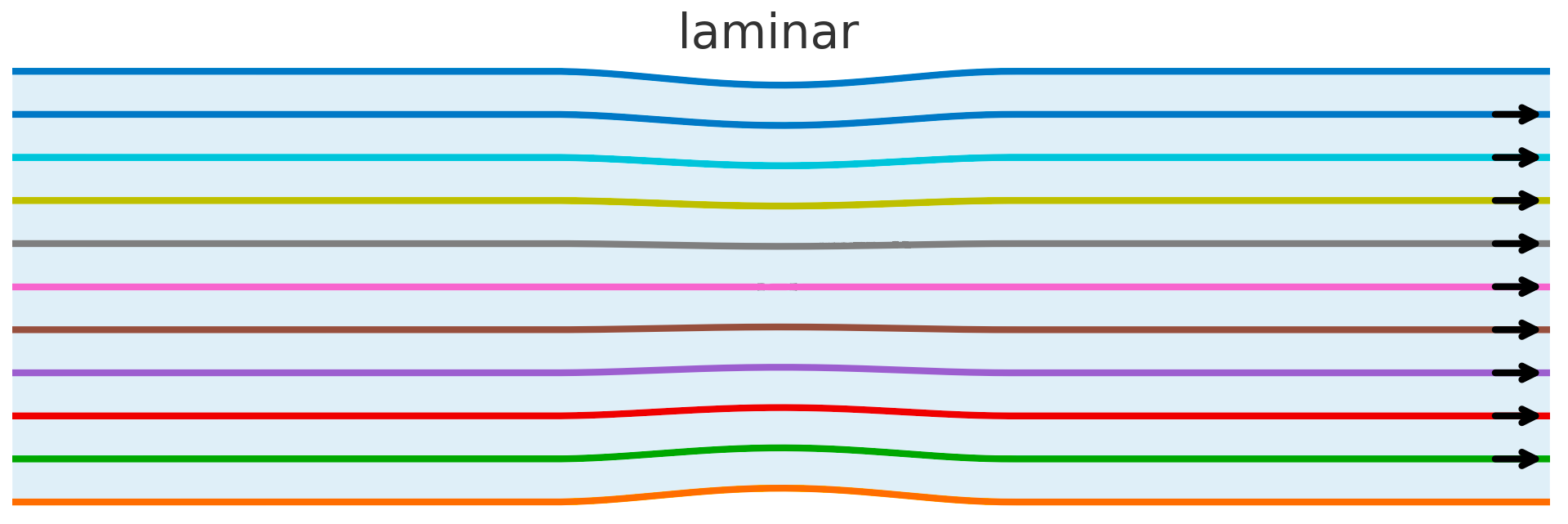}
      \par\vspace{0.4em}\small
      \textbf{Laminar flow.} \emph{Fluid:} viscous–dominated, low–Re regime; nearly parallel streamlines, negligible cross–stream mixing, no recirculation.\\
      \textbf{LLM Semantic Flow:} uniformly low spectral curvature $\kappa$, small steady $\Delta L$, and high alignment between the step and the belief push (steady refinement).
    \end{minipage}\hfill
    \begin{minipage}[t]{0.49\textwidth}\centering
      \includegraphics[width=\linewidth,height=\panelht,keepaspectratio]{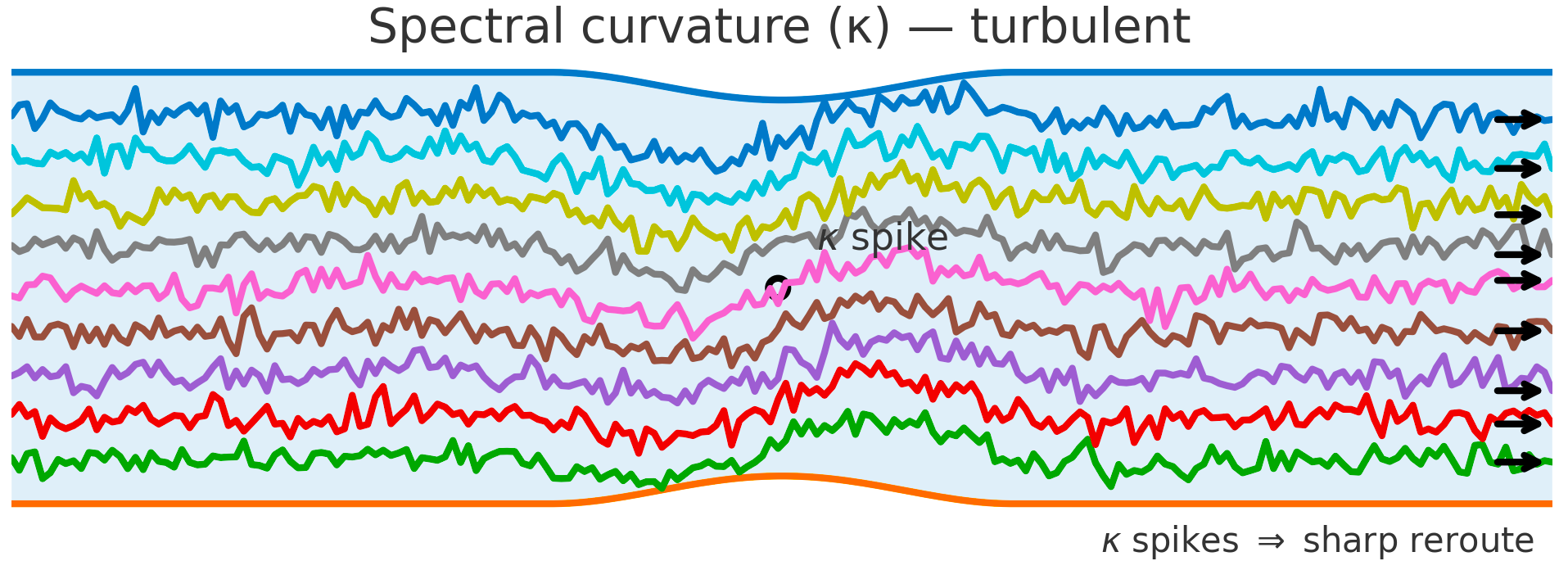}
      \par\vspace{0.4em}\small
      \textbf{Spectral curvature ($\kappa$) — turbulent.} \emph{Fluid:} a bend induces sharp turning, higher shear, possible separation. \\
      \textbf{LLM Semantic Flow:} a localized $\kappa$ spike at the turning point marks a sharp reroute in representation space; quasi–linear segments before/after indicate a discrete semantic pivot (e.g., topic jump, shortcut, policy jolt).
    \end{minipage}%
  }

  \vspace{0.75em}

  \noindent\makebox[\textwidth][c]{%
    \begin{minipage}[t]{0.49\textwidth}\centering
      \includegraphics[width=\linewidth,height=\panelht,keepaspectratio]{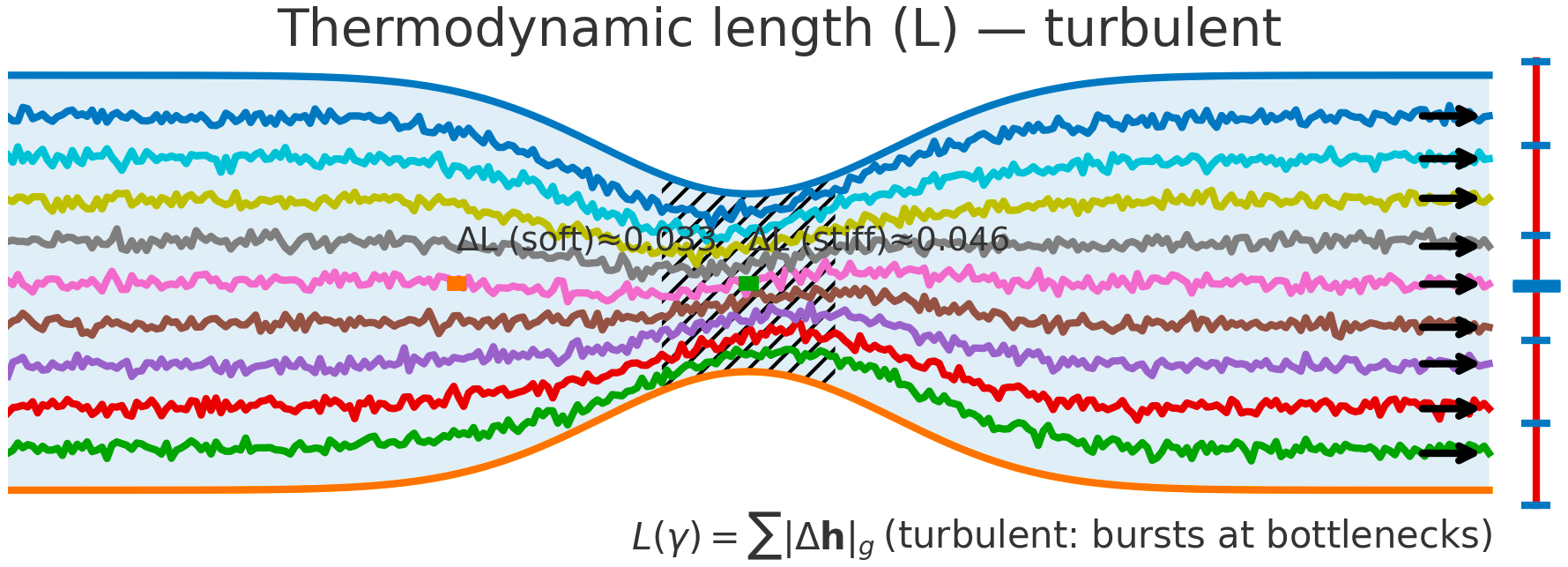}
      \par\vspace{0.4em}\small
      \textbf{Thermodynamic length ($L$).} \emph{Fluid:} a constriction raises shear and pressure drop; energy dissipates fastest in the throat. \\
      \textbf{LLM Semantic Flow:} a stiffer metric band (hatched) and a rise in $\Delta L$ reveal a bottleneck where extra \emph{semantic effort} is paid to reshape belief (friction, detours, boundary crossing).
    \end{minipage}\hfill
    \begin{minipage}[t]{0.49\textwidth}\centering
      \includegraphics[width=\linewidth,height=\panelht,keepaspectratio]{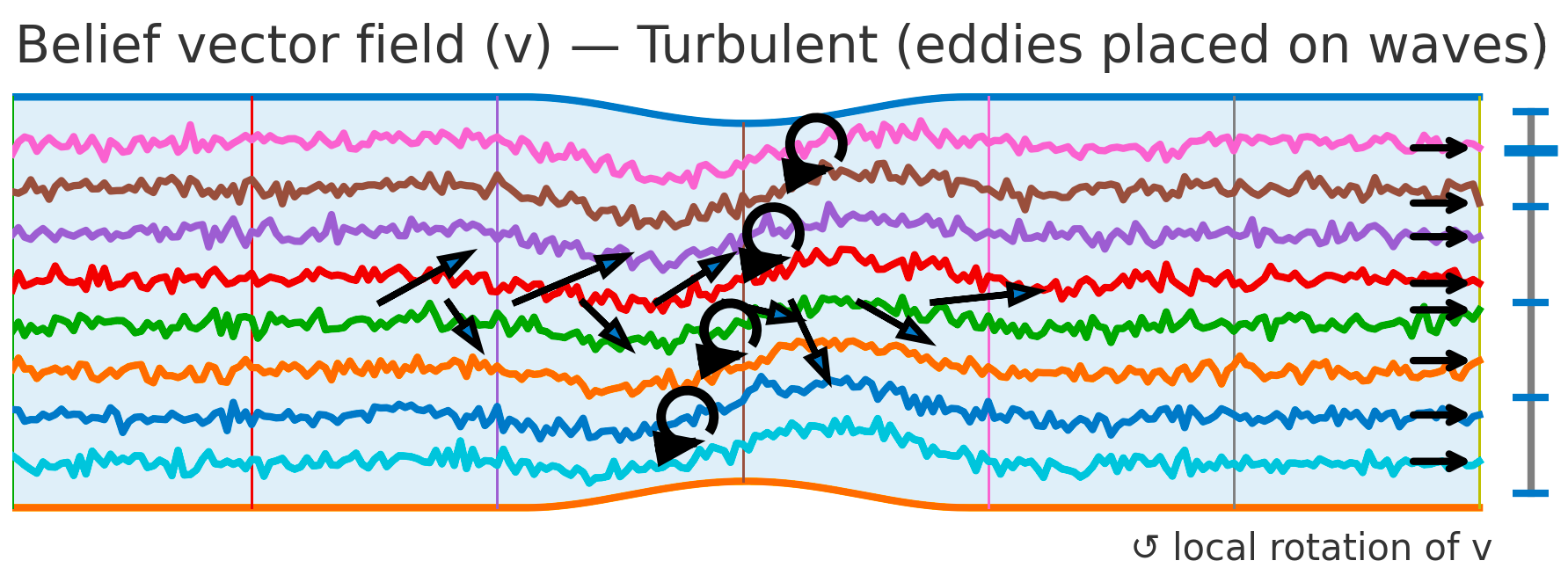}
      \par\vspace{0.4em}\small
      \textbf{Belief field ($\mathbf{v}$).} \emph{Fluid:} the velocity field sets transport; eddies (local curl) mark recirculation; alignment with streamlines indicates efficient conveyance. \\
      \textbf{LLM Semantic Flow:} $\mathbf{v}$ is the local push that most steeply changes the output law; longer arrows $\Rightarrow$ larger $\|\mathbf{v}\|$, and the side gauge shows $\cos\theta$ between $\mathbf{v}$ and the path tangent $\mathbf{T}$; circular loops on waves depict local recirculation that can trap or reinforce beliefs.
    \end{minipage}%
  }

  \caption{\textbf{LLM as an input$\to$output semantic channel.}
\emph{Model:} we read the forward pass as \emph{semantic hydrodynamics}—a prompt injects semantic mass that is transported through depth like a fluid through a shaped conduit.
\textbf{Bend} (\emph{top row}): curvature $\kappa$ distinguishes \emph{laminar} refinement from \emph{sharp} reroutes.
\textbf{Pay} (\emph{bottom left}): thermodynamic length $L$ localizes where effort concentrates via $\Delta L$ bursts (\emph{bottlenecks}).
\textbf{Push} (\emph{bottom right}): the belief field $\mathbf{v}$ reveals whether a layer update directly \emph{advances belief} (high alignment) or \emph{reorganizes information} (low alignment); eddies signal \emph{local recirculation}.\\
\textbf{Why this lens:} weight--space and attention views are \emph{non--identifiable} and unstable across checkpoints; nDNA instead reads the \emph{on--input trajectory} and its information geometry, yielding \emph{coordinate--free}, behavior--first measurements.\\
\textbf{Vision:} treat inner computation as a \emph{measurable flow} so that bends, effort, and push become quantifiable traits of cognition—comparable across inputs, layers, models, and training phases.\\
\textbf{Benefits:} \emph{actionable diagnostics}—$\kappa$ spikes flag brittle turns, $\Delta L$ bursts expose capacity bottlenecks, low $\cos\theta$ (between $\mathbf{v}$ and the tangent $\mathbf{T}$) indicates movement that does not immediately update belief; \emph{stable comparability}—geometry--based fingerprints are robust to neuron permutations and head--role drift; \emph{governance hooks}—set thresholds on $\kappa$ or $\Delta L$, track fingerprint drift after fine--tuning/pruning, and audit capacity before release.}
  \label{fig:ndna_semantic_flow_minipage}
\end{figure*}

\medskip
\noindent
\subsubsection{Why semantic hydrodynamics matters - (deeper intuition)}
\begin{itemize}[leftmargin=*,itemsep=0.4em]
  \item \textbf{We govern \emph{behavior}, not coordinates.} Operational concerns---\emph{robustness, safety, bias, faithfulness}---attach to what the model \textbf{does} on an input, not to how its weights are labeled. Two checkpoints can behave the same while their parameters and attention differ. In short: \textbf{the weights are the map; the trajectory is the territory}.
  \item \textbf{Invariance beats introspection.} Coordinate-bound stories change under neuron permutations, subspace rotations, or low-rank refactorings; the \textbf{path an input carves} and its \textbf{geometry} (length, curvature, alignment) are \textbf{invariant} because they are measured by \textbf{how predictions would change}, not by which index moved.
  \item \textbf{Geometry turns cognition into observables.} An information metric acts as local \textbf{stiffness}: soft directions barely affect the output; stiff directions swing the predictive law. With that ruler, we quantify \textbf{how far} the model travels to reshape belief (thermodynamic length $L$), \textbf{where} it turns its internal argument (spectral curvature $\kappa$), and \textbf{what} pushes change locally (belief field $\mathbf{v}$) \citep{sivak2012thermodynamic,hyvarinen2005estimation}.
  \item \textbf{The hydrodynamics metaphor is operational.} Like fluid in a channel, semantic flow shows \textbf{corners, constrictions, and eddies}: sharp bends $\Rightarrow$ high $\kappa$; narrow throats $\Rightarrow$ bursts in $\Delta L$; local recirculation $\Rightarrow$ rotational structure in $\mathbf{v}$. These are \textbf{measurable}, per-layer signals on the actual computation.
\end{itemize}

\medskip
\noindent\textbf{\emph{What this buys us} (concrete payoffs).}
\begin{itemize}[leftmargin=*,itemsep=0.4em]
  \item \textbf{Behavior-first invariance.} Reading $\kappa$, $L$, and $\mathbf{v}$ on the trajectory yields \textbf{fingerprints} that are \textbf{comparable} across models, seeds, and checkpoints---even when weights or head roles reshuffle.
  \item \textbf{Local diagnostics.} \textbf{$\kappa$ spikes} flag brittle decision pivots; \textbf{$\Delta L$ bursts} expose capacity bottlenecks or lossy transformations; \textbf{low alignment} (small $\cos\theta$ between $\mathbf{v}$ and the tangent $\mathbf{T}$) marks layers that \textbf{move without updating belief} (staging or detours).
  \item \textbf{Governance hooks.} \textbf{Geometry budgets and thresholds}---max $\kappa$, allowable $\Delta L$ per slice, minimum alignment---become \textbf{pre-release gates}; nDNA fingerprints support \textbf{drift monitoring} after fine-tuning, pruning, quantization, or alignment.
  \item \textbf{Comparative forensics.} Because $\kappa/L/\mathbf{v}$ are tied to the output law, we can \textbf{attribute performance deltas} to \textbf{where in depth} the flow changed (e.g., a new bend from fine-tuning, an effort spike from quantization) instead of to unstable weight indices.
\end{itemize}

\medskip
\noindent\textbf{\emph{Rule of thumb}.}
If the goal is to \textbf{explain}, \textbf{compare}, or \textbf{govern} deployed behavior, analyze the \textbf{flow of meaning} that the input actually experiences. In nDNA: \textbf{curvature} says \emph{where it bends}, \textbf{thermodynamic length} says \emph{how much it pays}, and the \textbf{belief field} says \emph{what pushes it}---all with a ruler calibrated to the model's own predictions.

We further posit that \textbf{cultural provenance} induces a distinct \emph{layerwise calibration effect}, predominantly localized in the final decoder layers $\ell \in [20, 30]$, where sociolinguistic priors exert the strongest influence on output distribution. To capture this, we introduce the \textbf{nDNA Score}--a composite diagnostic unifying: \textbf{(i)} \emph{Spectral curvature} $\kappa_\ell$, reflecting the compression and warping of conceptual flow; \textbf{(ii)} \emph{thermodynamic length} $\mathcal{L}_\ell$, quantifying the epistemic effort required to traverse belief transitions; and \textbf{(iii)} the norm of the \emph{Belief Vector Field} $\|\mathbf{v}_\ell^{(c)}\|$, measuring the directional intensity of latent cultural drift.

Together, these dimensions form a latent semantic fingerprint--a high-dimensional, biologically inspired signature of internal cognition--enabling us to \textbf{trace}, \textbf{compare}, and \textbf{govern} the \emph{neural evolution} of foundation models with unprecedented granularity.

\subsection{Spectral Curvature ($\kappa_\ell$): A Geometric Lens on Latent Bending}

\noindent
\textbf{What is spectral curvature?} In classical geometry, curvature quantifies how much a path deviates from being straight--measuring local bending of a trajectory. In \textbf{spectral geometry} and \textbf{harmonic analysis}, curvature extends to how signals or paths behave in frequency space or under operators that encode structure (e.g., Laplacians, difference operators). \emph{Spectral curvature} refers to curvature derived through such operators--capturing the \emph{shape of latent signals} as they evolve across layers of a model.

\vspace{0.8em}

\noindent
\textbf{Why spectral for latent manifolds?}  
In foundation models, hidden representations form a sequence of activations $\{ h_\ell \}_{\ell=0}^L$ across layers. These representations trace a path in high-dimensional latent space. The \emph{shape} of this path encodes the model’s \textbf{internal conceptual flow}--how its beliefs evolve as it integrates priors, inputs, and alignment constraints. \textbf{Spectral operators} (such as discrete Laplacians or difference operators) naturally quantify how this path bends or accelerates--making them ideal for probing internal geometry. Unlike mere distance measures, \emph{spectral curvature} reflects \textbf{intrinsic shape}, invariant under reparameterization.

\vspace{0.8em}

\noindent
\textbf{Formulation and derivation.}  
Consider hidden activations $h_\ell \in \mathbb{R}^d$ at each layer $\ell$. The \textbf{first-order difference}
\[
\Delta h_\ell := h_\ell - h_{\ell-1}
\]
approximates the local directional change of latent states--a discrete analogue of \emph{velocity} in latent space.

\vspace{0.5em}

\noindent
To capture bending, we compute the change in this directional flow--the \textbf{second-order difference}:
\[
\Delta^2 h_\ell := \Delta h_{\ell+1} - \Delta h_\ell = (h_{\ell+1} - h_\ell) - (h_\ell - h_{\ell-1}) = h_{\ell+1} - 2 h_\ell + h_{\ell-1}.
\]
This operator acts like a \emph{discrete Laplacian} along the latent path, highlighting where the model’s \textbf{internal belief flow} deviates from a straight trajectory.

\vspace{0.5em}

\begin{tcolorbox}[
  colback=gray!5!white,
  colframe=black!75!white,
  boxrule=0.5pt,
  arc=2pt
]
\centering
\emph{Spectral curvature at layer $\ell$ is defined as:}
\[
\kappa_\ell := \big\| \Delta^2 h_\ell \big\| = \big\| h_{\ell+1} - 2 h_\ell + h_{\ell-1} \big\|
\]
\end{tcolorbox}

\vspace{0.5em}

\noindent
In continuous form, this corresponds to:
\[
\kappa(s) = \left\| \frac{d^2 h(s)}{ds^2} \right\|
\]
where $s$ parameterizes depth through the network. Our discrete $\kappa_\ell$ provides a practical, layerwise estimator.

\vspace{0.8em}

\begin{figure}[H]
    \centering
    \includegraphics[width=0.9\linewidth]{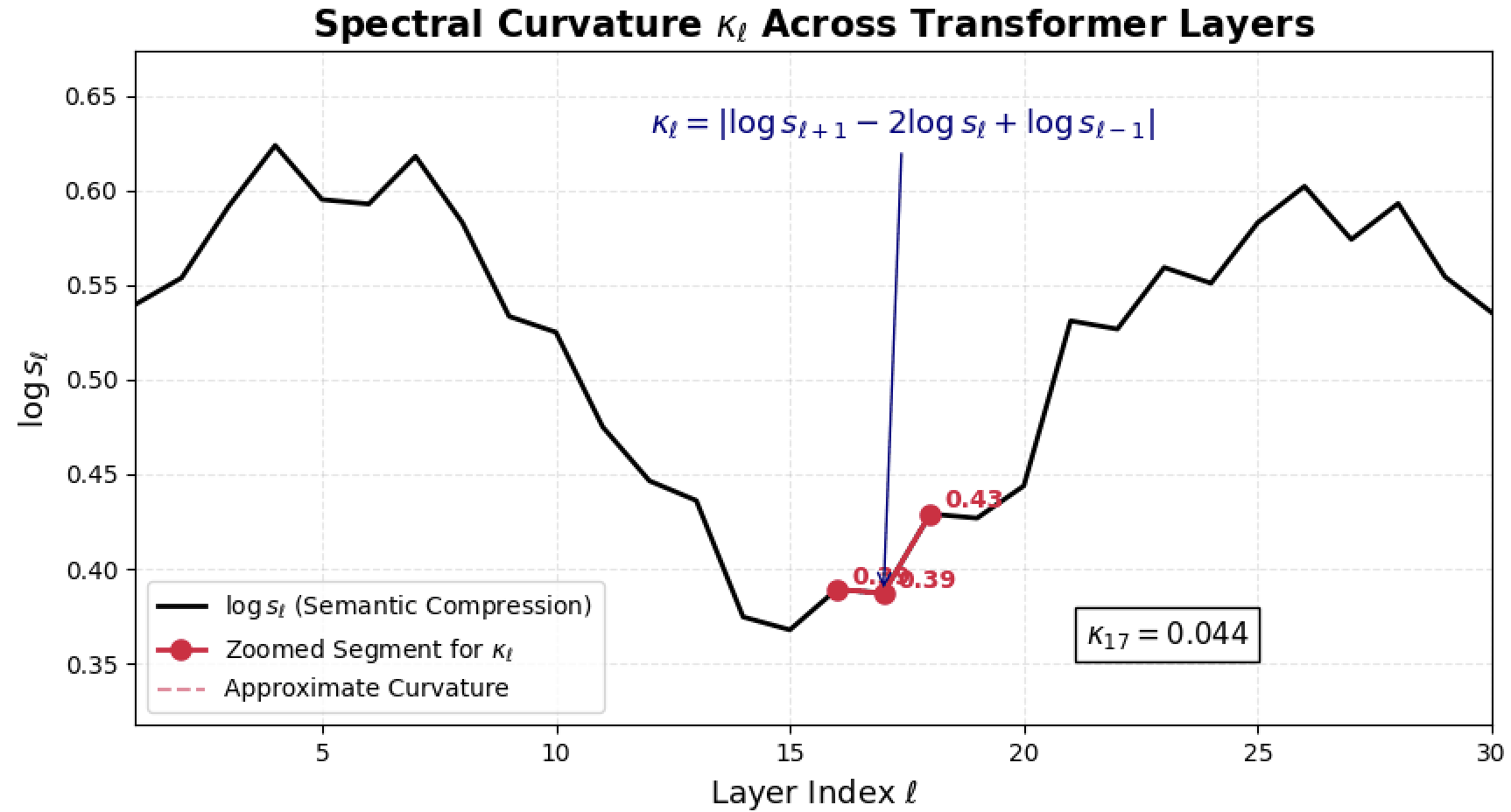}
    \includegraphics[width=0.9\linewidth]{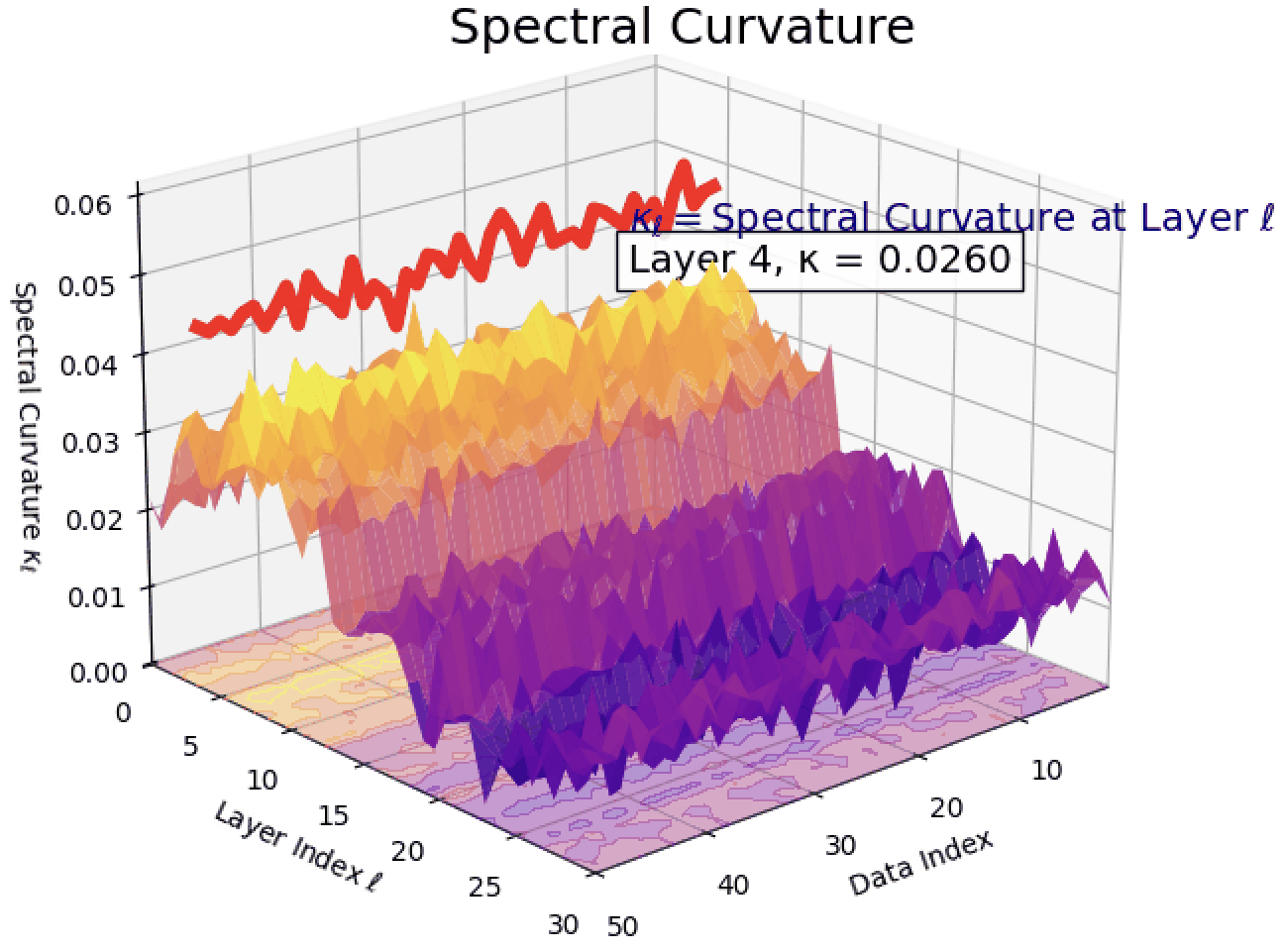}
    \vspace{-4mm}
    \caption{\textbf{Spectral Curvature} ($\boldsymbol{\kappa_\ell}$) quantifies second-order deviations in latent representations across transformer layers--computed via the discrete geometric operator $\boxed{\kappa_\ell := \| h_{\ell+1} - 2 h_\ell + h_{\ell-1} \|}$. High curvature signals \emph{semantic inflection points} where internal geometry bends sharply--often in \textbf{culturally dense}, \textbf{ideologically loaded}, or \textbf{epistemically volatile} regions. Peaks in $\kappa_\ell$ typically emerge in upper decoder layers ($\ell \in [21,30]$), where the model accommodates sociolinguistic priors during alignment, multicultural or multilingual fusion. Within the \textbf{\textit{n}}\textnormal{DNA} framework, such curvature reflects \emph{latent inheritance dynamics}, offering a fine-grained geometric fingerprint of representational restructuring.}
    \label{fig:spectral_curvature}
\end{figure}

\noindent
\textbf{Why is this meaningful?}  
Peaks in $\kappa_\ell$ mark layers where internal geometry is most dynamic--zones of \emph{semantic inflection}, \emph{belief compression}, or \emph{ideological absorption}. These are the structural signatures of \textbf{internal epistemic adaptation}, essential to trace cultural inheritance and alignment drift.

\vspace{0.3em}

\noindent
\textbf{Lineage and context.}  
Spectral curvature builds on tools from \textbf{geometric deep learning}, \textbf{equivariant architectures}, \textbf{Ricci flow in machine learning}, and \textbf{spectral graph analysis}~\citep{farzam2024ricci, cho2023mixedcurvature, gasteiger2021gemnet, xu2022spherical, konf2021hierarchical, ying2021equivariant, hu2022lie, hess2023spectral, wang2021geomtransformer, raposo2023spectral}. Within \textbf{\textit{n}}\textnormal{DNA}, it serves as a \textbf{principled geometric fingerprint}--revealing not only \emph{what} is encoded, but \emph{how} internal belief pathways are reshaped to encode it. Figure~\ref{fig:spectral_curvature} illustrates how spectral curvature \(\kappa_\ell\) measures second-order geometric changes in latent representations across transformer layers, revealing critical semantic inflection points that reflect nuanced, layer-specific restructuring in belief and ideologically influenced model epistemic.

\subsection{Thermodynamic Length ($\mathcal{L}_\ell$): Epistemic Effort Across Layers}

\noindent
\textbf{What is thermodynamic length?}  
In \textbf{statistical thermodynamics} and \textbf{information geometry}, \emph{thermodynamic length} measures the cumulative effort--or “\emph{work}”--required for a system to transition between states on a statistical manifold. It integrates local gradient energy along a trajectory, providing an \emph{intrinsic cost measure} that is independent of parametrization.

\vspace{0.5em}

\noindent
\textbf{Why thermodynamic length for foundation models?}  
In foundation models, layers trace a \textbf{path through latent belief space}. 
As input data and alignment priors reshape activations, the model expends internal \textbf{computational effort} to adjust its belief state. 
\emph{Thermodynamic length quantifies this latent effort} --- measuring not just \emph{what} the model knows, but \emph{how hard} it works to adapt that knowledge across layers in response to epistemic pressures (e.g., cultural fusion, alignment shifts).

\vspace{0.3em}

\noindent
\textbf{Mathematical intuition.}  
Let $h_\ell$ denote the latent state at layer $\ell$, and $\mathcal{M}$ the model’s latent manifold. 
Layer transitions define a curve 
$\gamma: [0,L] \to \mathcal{M}$ 
whose thermodynamic length is
\[
\boxed{
\mathcal{L}(\gamma) = \int_0^L \sqrt{ \big\langle \dot{\gamma}(s),\, \mathcal{G}_{\mathrm{Fisher}}\, \dot{\gamma}(s) \big\rangle }\, ds
}
\]
where $\mathcal{G}_{\mathrm{Fisher}}$ is the Fisher information metric. 
Here, $\mathcal{L}(\gamma)$ represents the \emph{intrinsic work} needed to traverse $\gamma$ on $\mathcal{M}$.

\vspace{0.3em}

\noindent
\textbf{Interpretation.}  
High thermodynamic length indicates regions where latent geometry \textbf{stretches} --- where the model’s belief space undergoes substantial reconfiguration to reconcile priors and input. 
This formalism reveals \emph{not just where} latent states change, but the \emph{cost structure of that change}. 
Zones of large $\mathcal{L}_\ell$ mark points of \textbf{alignment tension}, \textbf{cultural fusion}, or \textbf{complex reasoning}, where internal scaffolds are under maximum stress.

\vspace{0.3em}

\noindent
\emph{Thermodynamic length offers a window onto the model’s ``latent energy budget'' --- illuminating how internal belief states reshape to meet complexity, constraint, and context.}

\vspace{0.8em}

\noindent
\textbf{Formulation.}  
Let $p_\ell(y|x)$ denote the model’s conditional distribution at layer $\ell$ given input $x$. 
The local epistemic cost is reflected in the squared norm of the gradient of log-likelihood with respect to model parameters:
\[
\big\| \nabla_\theta \log p_\ell(x) \big\|^2.
\]
This quantity measures how much the model must \emph{adjust its parameters locally} at layer $\ell$ to improve its fit to input $x$. \emph{Thermodynamic length at layer $\ell$} aggregates this cost across the dataset $\mathcal{D}$:
\begin{tcolorbox}[
  colback=gray!5!white,
  colframe=black!75!white,
  boxrule=0.5pt,
  arc=2pt
]
\centering
\emph{Thermodynamic length at layer $\ell$ is defined as:}
\[
\mathcal{L}_\ell := \sum_{x \in \mathcal{D}} \big\| \nabla_\theta \log p_\ell(x) \big\|^2
= |\mathcal{D}| \, \mathbb{E}_{x \sim \mathcal{D}} \big\| \nabla_\theta \log p_\ell(x) \big\|^2.
\]
\end{tcolorbox}

\noindent
This formulation reveals that $\mathcal{L}_\ell$ captures both the \emph{average local effort} and its scaling with dataset size.  
Furthermore, in differential geometric terms, thermodynamic length can be written as a path energy:
\[
\mathcal{L}_\ell = \int_{\gamma_\ell} 
\left\langle \frac{d h_\ell}{ds}, 
\mathcal{G}_{\mathrm{Fisher}}(h_\ell)
\frac{d h_\ell}{ds} \right\rangle ds
\]
where $h_\ell$ denotes latent trajectories at layer $\ell$, $\mathcal{G}_{\mathrm{Fisher}}$ the Fisher information metric, and $s$ arc length along $\gamma_\ell$. Thus, $\mathcal{L}_\ell$ can be seen as an \emph{energy integral over the belief manifold} -- capturing how much ``\emph{heat}'' or computational work is generated to reconcile prior belief state with new input at depth $\ell$.

\begin{figure}[H]
    \centering
    \includegraphics[width=0.9\linewidth]{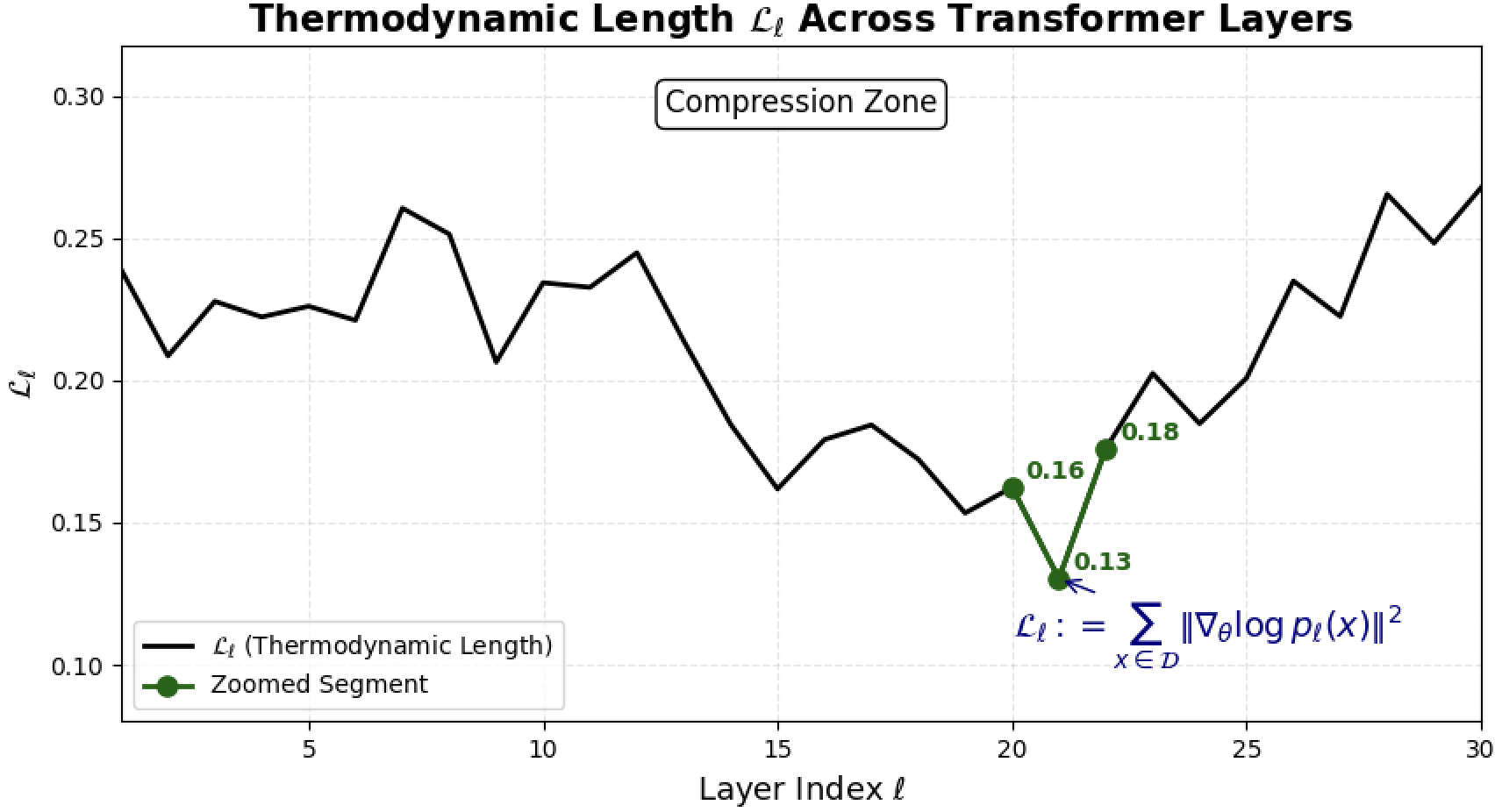}
    \includegraphics[width=0.9\linewidth]{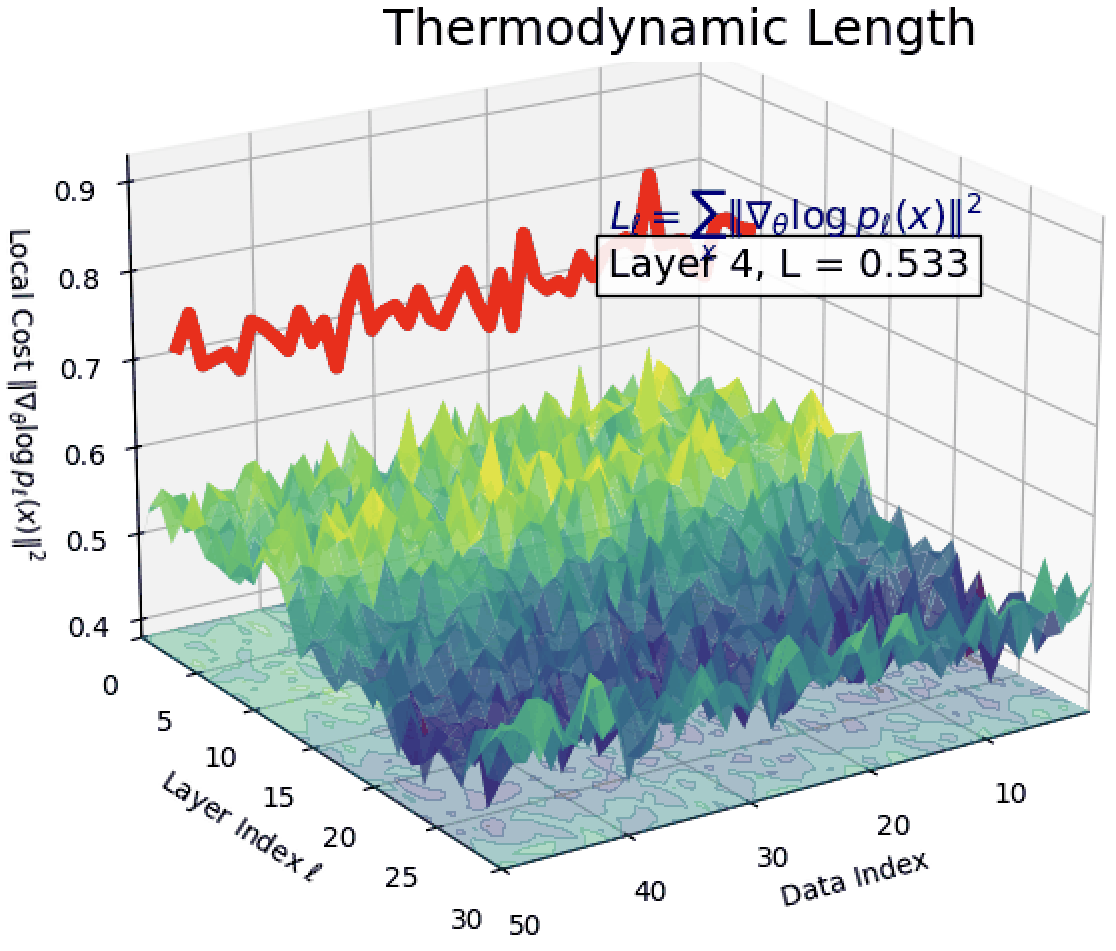}
    \vspace{-6mm}
    \caption{\textbf{Thermodynamic Length} $\boxed{\mathcal{L}_\ell := \sum_{x \in \mathcal{D}} \big\| \nabla_\theta \log p_\ell(x) \big\|^2}$ quantifies the \emph{epistemic work} performed across transformer layers, calculated as the \textbf{cumulative squared gradient norm of layerwise log-likelihoods}. Higher values signal \emph{internal resistance}--zones of significant restructuring, belief compression, or negotiation of conflicting priors. In culturally fine-tuned models, these peaks localize to upper decoder layers, indicating intense adaptation near output-generating blocks. Within the \textbf{\textit{n}}\textnormal{DNA} construct, $\mathcal{L}_\ell$ helps reveal latent epistemic effort that underlies surface-level behavior. This metric thus provides a nuanced window into where and how models internally allocate effort during learning and inference.}
    \label{fig:thermo_length}
\end{figure}

\vspace{0.5em}

\noindent
\textbf{Why is this meaningful?}  
Unlike static capacity metrics or weight magnitudes, $\mathcal{L}_\ell$ is \emph{dynamically grounded}: it measures where the model actively strains to reconcile competing epistemic demands. In regions of high $\mathcal{L}_\ell$, the model’s \textbf{latent geometry} is under tension--\emph{reshaping itself} to accommodate alignment constraints, cultural priors, or multilingual semantics.

\vspace{0.3em}

\noindent
\textbf{Lineage and context.}  
This diagnostic builds on the \textbf{Fisher–Rao metric} in \textbf{information geometry} and \textbf{thermodynamic length formalism} from statistical physics~\citep{crooks2007measuring, oliviero2023thermodynamics, farzam2024ricci, wagner2023thermodynamic}. Thus \textbf{\textit{n}}\textnormal{DNA} provides a \emph{complementary view} to spectral curvature--capturing not where the model bends, but \emph{how hard it works} to do so. Together, these axes form a \textbf{neurogeometric anatomy} of latent belief adaptation. 

Figure~\ref{fig:thermo_length} shows thermodynamic length \(\mathcal{L}_\ell\), quantifying latent epistemic effort and semantic restructuring across transformer layers.

\subsection{Belief Vector Field ($\mathbf{v}_\ell^{(c)}$): Cultural Drift in Latent Space}

\noindent
\textbf{What is the Belief Vector Field} --  
In \textbf{differential geometry} and \textbf{physics}, a \emph{vector field} describes a directional force applied at each point of a space. Inspired by this, the \textbf{Belief Vector Field} models the \emph{directional semantic force} that a specific culture or value system exerts on a model’s latent representations. It encodes \emph{where}, \emph{how strongly}, and \emph{in what direction} cultural priors act within the model’s internal geometry--functioning as a \textbf{semantic compass} through the latent manifold.

\begin{figure}[H]
    \centering
    \includegraphics[width=\linewidth]{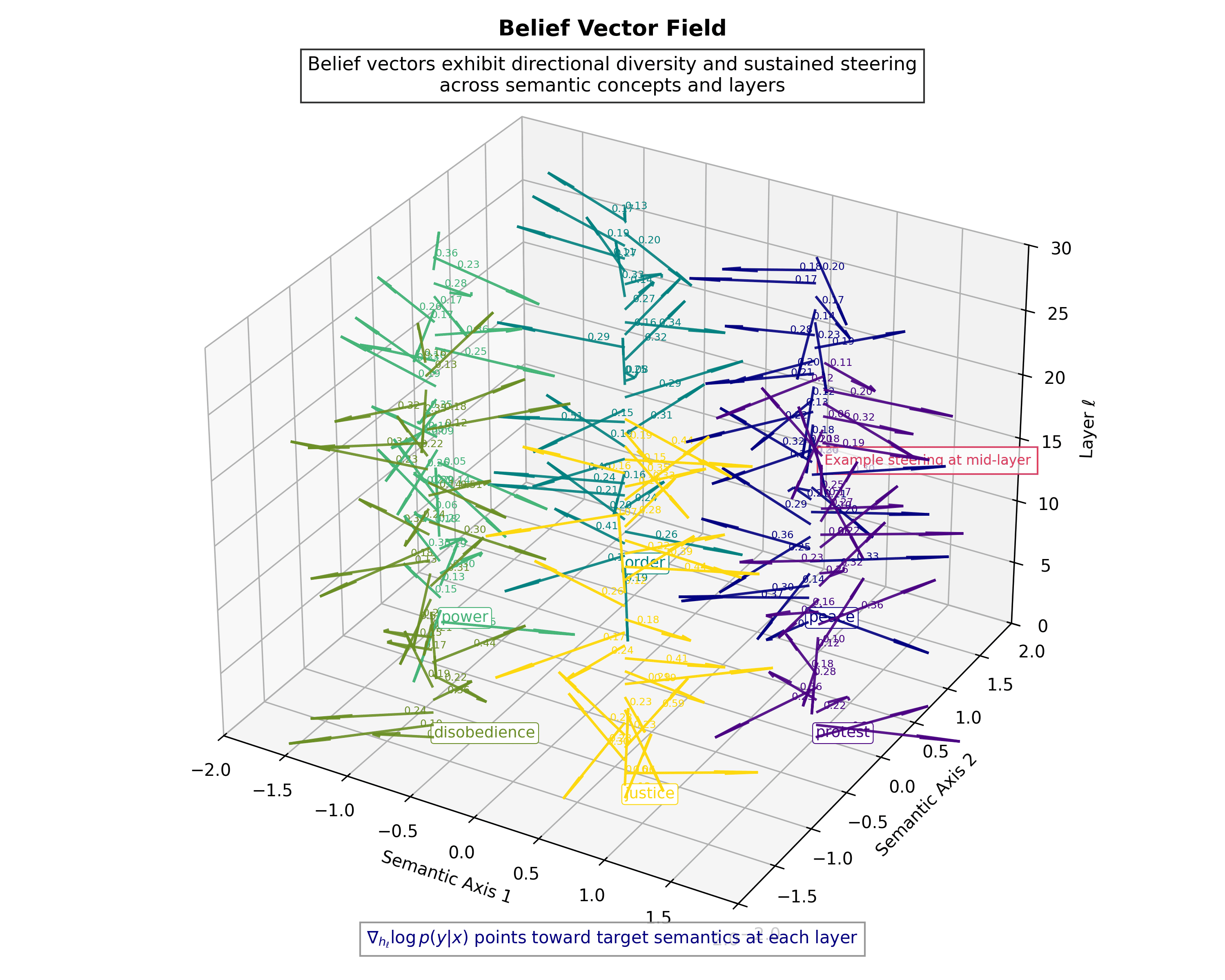}
    \vspace{-8mm}
    \caption{
    \textbf{Belief Vector Field Visualization}: 
    $\mathbf{v}_\ell^{(c)} = 
    \mathbb{E}_{x \sim \mathcal{P}^{(c)}_{\mathrm{CIVIC}}}
    \left[
    \nabla_{h_\ell} \log p(y \mid x)
    \right]$
    represents the \emph{belief semantic steering force} at layer~$\ell$ toward concept~$c$, conditioned on CIVIC cultural priors (cf.~\cref{sec:aether_benchmark}).
    \textbf{Large magnitudes} (e.g., $\| \mathbf{v}_\ell^{(c)} \| \in [0.15, 0.50]$) indicate \emph{strong directional pressure}--zones where cultural values actively reshape latent geometry.
    \textit{Color-coded arrows} trace distinct conceptual trajectories (\textcolor[rgb]{0.29,0.00,0.51}{protest}, \textcolor[rgb]{0.00,0.00,0.50}{peace}, \textcolor[rgb]{0.00,0.50,0.50}{order}, \textcolor[rgb]{0.24,0.70,0.44}{power}, \textcolor[rgb]{0.42,0.56,0.14}{disobedience}, \textcolor[rgb]{0.85,0.65,0.13}{justice}), while numeric labels quantify local steering strength.
    Upper layers ($\ell \ge 20$) typically exhibit \textbf{epistemic reorientation}, where cultural priors most heavily influence belief encoding.
    Such visualizations reveal whether a model internalizes culturally contingent reasoning or merely mimics alignment at the output surface.
    }
    \label{fig:belief_vector_field}
\end{figure}

\vspace{0.5em}

\noindent
\textbf{Why a vector field for cultural influence?}  
While \textbf{spectral curvature} ($\kappa_\ell$) captures how sharply latent paths bend, and \textbf{thermodynamic length} ($\mathcal{L}_\ell$) how hard the model works during adaptation, neither tells us the \emph{source}, \emph{direction}, or \emph{origin} of that adaptation. The Belief Vector Field offers this missing piece: it traces the latent steering aka torison applied by culture-conditioned priors--\emph{where the model is being pushed in latent space, by what epistemic force, and toward which semantic direction}. This makes it a critical diagnostic for studying \textbf{cultural drift}, \textbf{ideological imprinting}, and \textbf{alignment tension}.

\noindent
\textbf{Formulation and derivation.}  
Let $p(y|x)$ denote the model’s conditional output distribution for input $x$, and let $h_\ell$ be the latent representation at layer $\ell$. The local belief gradient, $\nabla_{h_\ell} \log p(y|x)$, measures how a small change in $h_\ell$ would affect output confidence--a proxy for \emph{semantic force} at that layer. To extract the culturally conditioned semantic force, we compute its expectation over a culture-specific distribution $\mathcal{P}^{(c)}$:
\begin{tcolorbox}[
  colback=gray!5!white,
  colframe=black!75!white,
  boxrule=0.5pt,
  arc=2pt
]
\centering
\emph{Belief vector field at layer $\ell$ for a given manifold condition is defined as:}
\[
\mathbf{v}_\ell^{(c)} := \mathbb{E}_{x \sim \mathcal{P}^{(c)}} \left[ \nabla_{h_\ell} \log p(y|x) \right]
\]
\end{tcolorbox}
where $\mathcal{P}^{(c)}$ represents inputs emblematic of givem manifold condition $c$ (e.g., regional, linguistic, ideological contexts). This formulation captures not just latent deformation, but \emph{its cause}: how cultural priors exert directional influence within the belief manifold.

\noindent
\textbf{Why is this meaningful?}  
$\mathbf{v}_\ell^{(c)}$ provides a directional lens on latent dynamics. High $\| \mathbf{v}_\ell^{(c)} \|$ signals regions where the model is \emph{actively redirected} by external cultural forces--offering diagnostic power for detecting \textbf{ideological drift}, \textbf{semantic conflict}, or \textbf{bias inheritance}. Unlike $\kappa_\ell$ or $\mathcal{L}_\ell$, which capture internal geometry, $\mathbf{v}_\ell^{(c)}$ reveals \emph{external epistemic pressure} and its directional impact.

\vspace{0.3em}

\noindent
\textbf{Lineage and context.}  
This diagnostic builds upon belief geometry, alignment drift studies, and cultural bias tracing in NLP~\citep{wang2023culturalbias, zhou2023alignmentdrift, shen2023beliefgeometry, arora2023stereoset, bommasani2023foundation, peng2024cultural, laurens2024anthropic, kang2024biasfairness, de2023latentbias, gao2023value}. Within the \textbf{\textit{n}}\textnormal{DNA} construct, it integrates with curvature and length to offer a holistic neurogeometric portrait--revealing \emph{how}, \emph{why}, and \emph{where} foundation models inherit, adapt, or distort beliefs under cultural influence.

\vspace{0.3em}

\noindent
\textbf{Interpretability in practice.}  
By mapping $\mathbf{v}_\ell^{(c)}$ across layers and cultures, we can trace \textbf{cultural provenance}, identify \textbf{ideological pressure zones}, and diagnose \textbf{inheritance asymmetry} in multilingual or aligned models. This directional fingerprint informs audits of model bias, robustness, and alignment integrity--providing the missing vectorial dimension in understanding machine cognition.

\subsection{\textbf{\textit{n}}\textnormal{DNA}: Unified Epistemic Inheritance Measure}

\noindent
\textbf{Why a unified score?}  
While \textbf{spectral curvature} ($\kappa_\ell$), \textbf{thermodynamic length} ($\mathcal{L}_\ell$), and the \textbf{belief vector field norm} ($\| \mathbf{v}_\ell^{(c)} \|$) each offer unique insight into latent dynamics, they operate on distinct facets of \emph{epistemic geometry}:

\vspace{0.8em}

\begin{tcolorbox}[
  colback=blue!3!white,
  colframe=black!80!white,
  boxrule=0.6pt,
  arc=3pt,
  width=\textwidth
]
\footnotesize
\textbf{The nDNA score is a cumulative measure of latent geometry, quantifying how a large language model adapts its internal scaffolding to a given corpus.} 
It integrates three key components at each layer $\ell$:
\begin{itemize}[leftmargin=1.5em]
  \item \textbf{Curvature} ($\kappa_\ell$): how \emph{twisted} or \emph{bent} the latent manifold is; captures \emph{how sharply} internal trajectories bend -- a scalar measure of latent acceleration.
  \item \textbf{Length} ($\mathcal{L}_\ell$): how much \emph{latent work} or displacement occurs as representations evolve; quantifies \emph{how hard} the model works to adapt its beliefs -- a scalar effort integral.
  \item \textbf{Belief vector norm} ($\|\mathbf{v}_\ell^{(c)}\|$): how \emph{strong} the model’s belief signal is for that corpus; encodes \emph{where} and \emph{how strongly} cultural priors steer latent space -- a scalar magnitude derived from the vector field.
\end{itemize}
\begin{center}
\emph{Formally, we define the \textbf{\textit{n}}\textnormal{DNA} score as:}
\[\boxed{
\texttt{nDNA} := \sum_{\ell=1}^{L} \omega_\ell \cdot \kappa_\ell \cdot \mathcal{L}_\ell \cdot \| \mathbf{v}_\ell^{(c)} \|}
\]
\end{center}
\end{tcolorbox}

\vspace{0.5em}

\begin{figure*}[ht!]
\centering
\includegraphics[width=0.98\textwidth]{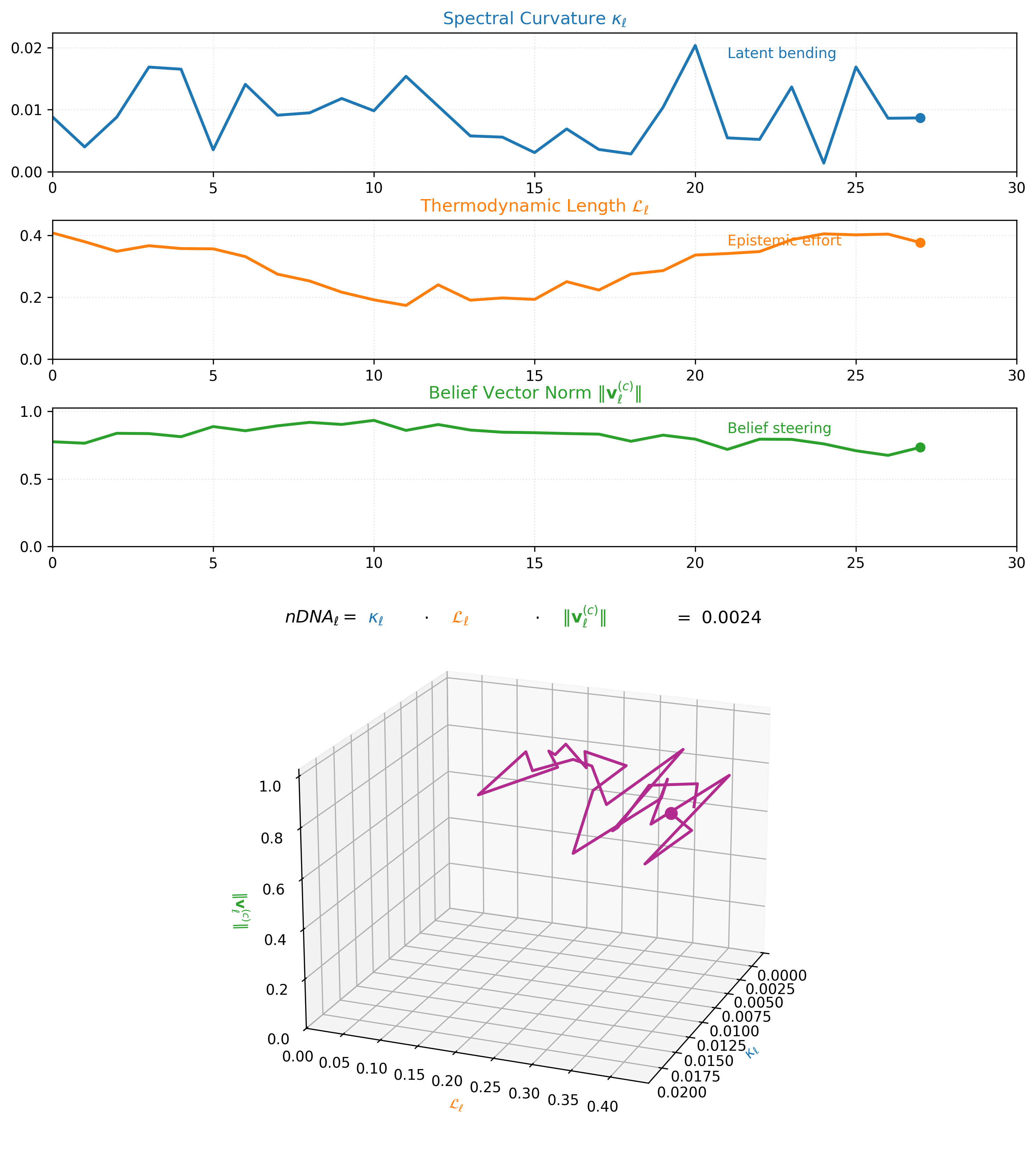}
\vspace{-4mm}
\caption{
\textbf{The compositional anatomy of neural DNA (\emph{n}DNA) through curvature, length, and belief geometry.}
This figure illustrates how \emph{n}DNA arises as a layered product of three latent quantities. 
First, \textbf{spectral curvature} $\boldsymbol{\kappa_\ell}$ measures latent manifold bending and flexibility (latent acceleration), indicating how sharply the internal geometry twists at layer $\ell$. 
Second, \textbf{thermodynamic length} $\boldsymbol{\mathcal{L}_\ell}$ quantifies the accumulated epistemic effort (latent adaptation energy) and reflects how hard the model works to reconcile prior beliefs with new input and alignment signals. 
Third, \textbf{belief vector norm} $\|\mathbf{v}_\ell^{(c)}\|$ encodes the magnitude of latent directional force imposed by corpus priors or alignment signals. 
The joint trajectory in $(\kappa_\ell, \mathcal{L}_\ell, \|\mathbf{v}_\ell^{(c)}\|)$ space, color-coded by the composite score, shows how bending, effort, and steering co-evolve across layers. 
The combined latent signature is formalized as $\mathit{nDNA}_\ell = \kappa_\ell \cdot \mathcal{L}_\ell \cdot \|\mathbf{v}_\ell^{(c)}\| = 0.0024$ (example layer), with high values identifying zones of intense latent reconfiguration where geometry and adaptation forces align. 
Color-keyed descriptors (``Latent bending'', ``Epistemic effort'', ``Belief steering'') guide visual interpretation. 
The figure illustrates how large language models coordinate latent bending, effort, and steering to build a neurogeometric scaffold that adapts flexibly to task complexity while remaining anchored in a universal latent structure.
}
\label{fig:ndna_story}
\end{figure*}

\vspace{0.8em}

Individually, these measures illuminate latent strain, adaptation cost, and cultural pressure. But to assess \emph{inheritance as a whole} -- how traits propagate through \textbf{fine-tuning}, \textbf{merging}, or \textbf{distillation} -- we must integrate these into a single diagnostic that reflects combined latent geometry, epistemic work, and directional influence.

\noindent
\textbf{Designing the composite measure.}  
Since $\kappa_\ell$ and $\mathcal{L}_\ell$ are scalars, and $\| \mathbf{v}_\ell^{(c)} \|$ reduces directional drift to scalar magnitude, their product forms a natural joint measure of:
\emph{internal bending} ($\kappa_\ell$),
\emph{internal epistemic effort} ($\mathcal{L}_\ell$),
and \emph{external drift pressure} ($\| \mathbf{v}_\ell^{(c)} \|$).
To balance their contributions across depth, we introduce layer weights $\omega_\ell$, emphasizing semantically active or epistemically significant layers (e.g., $\omega_\ell$ higher in upper decoder blocks).

\noindent
This composite score integrates scalar and vector-derived diagnostics into a unified measure of \emph{epistemic inheritance} -- quantifying the latent structure and cultural traits a model carries forward from its neural ancestry.

\vspace{0.3em}

\noindent
\textbf{Rationale for multiplicative integration.}  
This form spotlights layers where latent paths bend sharply, belief adaptation incurs significant effort, and cultural or alignment pressures apply strong directional force. 
High scores identify zones of \emph{intense latent reconfiguration}, where internal dynamics and external pressures converge to reshape the model’s reasoning space.

\vspace{0.3em}

\noindent
\textbf{Role of $\omega_\ell$.}  
The weight $\omega_\ell$ serves as a lens to prioritize semantically expressive, epistemically active regions of the network. It may be set uniformly, hand-tuned, or optimized against alignment drift benchmarks, bias metrics, or interpretability objectives.

\vspace{0.3em}

\noindent
\textbf{Interpretability and utility.}  
The $\texttt{nDNA}$ score provides a compact fingerprint of model inheritance:
\begin{itemize}[leftmargin=1.5em]
    \item It enables direct comparison of parent and child models post \textbf{fine-tuning}, \textbf{merging}, or \textbf{distillation}.
    \item It highlights zones of \emph{semantic mutation}, \emph{ideological absorption}, or \emph{cultural drift}.
    \item It serves as a proxy for \emph{latent epistemic integrity} -- quantifying the hidden cost and directionality of neural evolution.
\end{itemize}

\vspace{0.3em}

\noindent
\textbf{Conviction.}  
By unifying \textbf{spectral}, \textbf{thermodynamic}, and \textbf{vectorial} diagnostics, the $\texttt{nDNA}$ score functions as a \textbf{heritable geometry index} -- diagnosing how latent traits persist, mutate, or degrade as foundation models evolve.

\subsection{nDNA Geometry - A closer Look}

\noindent
The notion of \textbf{nDNA} arises from a simple yet profound insight: modern foundation models do not merely produce outputs--they embody a latent cognitive structure that governs how they reason, adapt, and evolve~\citep{bommasani2023foundation, ganguli2023reducing}. This latent structure is not directly encoded in model weights or activations alone; rather, it emerges in the internal geometry of belief formation, semantic flow, and epistemic adaptation across layers~\citep{liu2023hidden, wang2021geomtransformer}. We define the \textbf{nDNA geometry} of a model as the joint distribution of its 
\textbf{spectral curvature} ($\boldsymbol{\kappa_\ell}$), 
\textbf{thermodynamic length} ($\boldsymbol{\mathcal{L}_\ell}$), and 
\textbf{belief vector field norm} ($\| \mathbf{v}_\ell^{(c)} \|$)
layer-by-layer. 
This triad forms a high-dimensional semantic fingerprint that encodes a model's \emph{inheritance stability}, 
\emph{alignment dynamics}, 
and \emph{cultural drift}---analogous to how biological DNA records heritable traits and mutations~\citep{shen2023beliefgeometry, bakker2024uniting}.

\noindent
Table~\ref{tab:ndna_example} provides an \emph{illustrative example of nDNA geometry}, highlighting how these quantities vary across depth in a representative model. 
Rather than simple monotonic trends, we observe intricate layer-wise patterns: certain layers exhibit elevated curvature ($\kappa_\ell > 0.06$), signaling sharp latent reorientation~\citep{cho2023mixedcurvature}, while others concentrate thermodynamic length ($\mathcal{L}_\ell > 1.10$), reflecting zones of intense internal work to reconcile competing priors~\citep{crooks2007measuring, oliviero2023thermodynamics}. 
The belief vector norm $\| \mathbf{v}_\ell^{(c)} \|$ exposes the directional cultural force acting on the latent manifold~\citep{peng2024cultural, zhou2023alignmentdrift}, marking layers where external alignment or sociolinguistic conditioning exerts greatest influence. 
Together, these values form a geometry-specific trace that distinguishes models by their latent adaptation history.

\begin{table}[H]
\centering
\caption{
An \textbf{illustrative nDNA example}: that captures the \emph{semantic genome} of a foundation model through the joint interplay of \textbf{spectral curvature} ($\boldsymbol{\kappa_\ell}$), \textbf{thermodynamic length} ($\boldsymbol{\mathcal{L}_\ell}$), \textbf{belief vector norm} ($\| \mathbf{v}_\ell^{(c)} \|$) across layers. Each of these quantities offers a distinct geometric and epistemic lens: $\boldsymbol{\kappa_\ell}$ measures the \emph{local acceleration} of latent representations, $\boldsymbol{\mathcal{L}_\ell}$ quantifies the cumulative \emph{internal work} required to traverse the belief manifold, while $\| \mathbf{v}_\ell^{(c)} \|$ encodes the \emph{magnitude of cultural drift} imposed on latent activations. The \emph{color intensities} shown alongside each value reflect relative magnitude within column-specific ranges: \textcolor[rgb]{0,0.6,0}{\rule{1em}{1em}} low, 
\textcolor[rgb]{0.8,0.8,0}{\rule{1em}{1em}} moderate, \textcolor[rgb]{1,0.5,0}{\rule{1em}{1em}} high, \textcolor[rgb]{0.8,0,0}{\rule{1em}{1em}} very high. For this example, spectral curvature spans $\boldsymbol{\kappa_\ell} \in [0.0400, 0.0700]$, thermodynamic length $\boldsymbol{\mathcal{L}_\ell} \in [0.80, 1.20]$, and belief vector norm $\| \mathbf{v}_\ell^{(c)} \| \in [0.55, 0.75]$--revealing regions where the \emph{latent manifold bends}, \emph{epistemic energy intensifies}, or \emph{external priors steer internal cognition}. This triad forms what we term the model's \textbf{nDNA}: a compact, high-dimensional \emph{semantic fingerprint} 
that encodes the hidden geometry of belief. 
It enables us to diagnose zones of \emph{inheritance stability}, detect \emph{ideological absorption}, and trace \emph{latent mutations} introduced by fine-tuning, alignment, or architectural choice. The pattern of these quantities across layers constitutes a signature as unique as a biological genome -- a map of how artificial cognition evolves, 
remembers, and adapts.
}
\begin{tabular}{|c|c|c|c|l|}
\hline
\textbf{Layer} 
& $\boldsymbol{\kappa_\ell}$ 
& $\boldsymbol{\mathcal{L}_\ell}$ 
& $\|\mathbf{v}_\ell^{(c)}\|$ 
& \textbf{Belief Vector} $\mathbf{v}_\ell^{(c)}$ \\
\hline
20 & \cellcolor{green!20}0.0412 & \cellcolor{yellow!30}0.9123 & \cellcolor{orange!20}0.6521 & $[0.1204, -0.0502, 0.0896, \ldots, 0.0402]$ \\
21 & \cellcolor{green!30}0.0458 & \cellcolor{green!15}0.8123 & \cellcolor{red!30}0.7523 & $[0.1301, -0.0351, 0.0950, \ldots, 0.0431]$ \\
22 & \cellcolor{yellow!20}0.0523 & \cellcolor{orange!30}1.0120 & \cellcolor{green!20}0.5823 & $[0.1423, -0.0312, 0.0994, \ldots, 0.0488]$ \\
23 & \cellcolor{orange!20}0.0581 & \cellcolor{yellow!20}0.9021 & \cellcolor{yellow!30}0.6912 & $[0.1534, 0.0270, 0.1042, \ldots, 0.0512]$ \\
24 & \cellcolor{red!20}0.0639 & \cellcolor{red!40}1.1023 & \cellcolor{green!15}0.5520 & $[0.1667, 0.0205, 0.1105, \ldots, 0.0543]$ \\
25 & \cellcolor{yellow!30}0.0505 & \cellcolor{orange!20}0.9420 & \cellcolor{red!40}0.8124 & $[0.1602, -0.0251, 0.1081, \ldots, 0.0504]$ \\
26 & \cellcolor{green!15}0.0398 & \cellcolor{green!20}0.8520 & \cellcolor{yellow!20}0.6120 & $[0.1251, 0.0450, 0.0912, \ldots, 0.0418]$ \\
27 & \cellcolor{yellow!20}0.0512 & \cellcolor{red!30}1.0520 & \cellcolor{orange!30}0.7222 & $[0.1455, -0.0322, 0.1005, \ldots, 0.0477]$ \\
28 & \cellcolor{orange!30}0.0590 & \cellcolor{yellow!30}0.9320 & \cellcolor{green!30}0.5721 & $[0.1577, 0.0285, 0.1078, \ldots, 0.0499]$ \\
29 & \cellcolor{red!30}0.0672 & \cellcolor{orange!30}1.0123 & \cellcolor{yellow!20}0.6322 & $[0.1701, -0.0198, 0.1142, \ldots, 0.0533]$ \\
30 & \cellcolor{orange!20}0.0555 & \cellcolor{green!15}0.8221 & \cellcolor{red!20}0.7720 & $[0.1620, -0.0242, 0.1101, \ldots, 0.0510]$ \\
\hline
\end{tabular}
\label{tab:ndna_example}
\end{table}

\section{The Corpus Dependence of nDNA: A Necessary Feature, Not a Flaw}

\begin{figure}[ht!]
\centering
\begin{tikzpicture}[node distance=1.2cm and 1.5cm, align=center, font=\small]

\node (dna_label) at (0,0.7) {\textbf{Universal DNA (shared genome)}};
\draw[decorate, decoration={coil, aspect=0.4, amplitude=4pt, segment length=6pt}, thick, blue] (0,0.5) -- (0,-0.5);
\node (dna) at (0,-0.5) {};

\node (methyl) [rectangle, draw, fill=gray!20] at (-3,-2) {DNA methylation \\ (suppression)};
\node (acetyl) [rectangle, draw, fill=green!20] at (3,-2) {Histone acetylation \\ (activation)};
\draw[->, thick, gray] (dna) -- (methyl);
\draw[->, thick, green!70!black] (dna) -- (acetyl);

\node (tf_neuron) [rectangle, draw, fill=green!10, below=1.5cm of methyl] {Neuron TFs \\ (NeuroD, REST)};
\node (tf_hep) [rectangle, draw, fill=orange!10, below=1.5cm of acetyl] {Hepatocyte TFs \\ (HNF4, C/EBP$\alpha$)};
\draw[->, thick] (methyl) -- (tf_neuron);
\draw[->, thick] (acetyl) -- (tf_hep);

\node (chrom_neuron) [rectangle, draw, dashed, below=1.2cm of tf_neuron] 
{Open chromatin \\ neuron genes on};
\node (chrom_hep) [rectangle, draw, dashed, below=1.2cm of tf_hep] 
{Compact chromatin \\ neuron genes off};
\draw[->] (tf_neuron) -- (chrom_neuron);
\draw[->] (tf_hep) -- (chrom_hep);

\node (gene_neuron) [rectangle, draw, rounded corners=3pt, fill=green!5, below=1cm of chrom_neuron] 
{Synaptic genes ON};
\node (gene_hep) [rectangle, draw, rounded corners=3pt, fill=orange!5, below=1cm of chrom_hep] 
{Detox/metabolic genes ON};
\draw[->] (chrom_neuron) -- (gene_neuron);
\draw[->] (chrom_hep) -- (gene_hep);

\node (func_neuron) [below=0.7cm of gene_neuron] {Function: signaling, plasticity};
\node (func_hep) [below=0.7cm of gene_hep] {Function: glucose metabolism, detox};

\end{tikzpicture}

\caption{
\textbf{A hierarchical view of universal DNA and context-sensitive gene expression, as a biological parallel to nDNA latent scaffolding in LLMs.}
This figure illustrates how the \emph{same genome} (depicted as a universal DNA helix at the top) produces distinct functional outcomes through a layered and structured regulatory architecture. 
The \textbf{first regulatory layer} consists of \textit{epigenetic modifications}, including DNA methylation (linked with gene silencing) and histone acetylation (linked with gene activation) 
\citep{bird2007perceptions, clapier2017mechanisms}. 
These modifications influence chromatin accessibility, setting the stage for context-specific transcriptional control.  
The \textbf{second layer} involves \textit{cell-type-specific transcription factors (TFs)} -- for example, NeuroD and REST in neurons, or HNF4 and C/EBP$\alpha$ in hepatocytes -- which bind regulatory DNA elements 
and integrate signaling cues to guide gene expression programs \citep{lambert2018human, davidson2006gene}.  
The \textbf{third layer} reflects the resultant chromatin state: open, transcriptionally permissive configurations in neurons for synaptic gene activation, versus compact, repressive configurations in hepatocytes where those genes are silent 
\citep{thurman2012accessible, dekker2013exploring}.  
Finally, this hierarchical regulatory control produces \textit{functionally specialized gene programs}: neurons activate synaptic plasticity and axon signaling genes; hepatocytes activate detoxification and glucose metabolism genes 
\citep{lewin2013genes, alon2006introduction}.  
This layered architecture provides a powerful biological analogy for \emph{nDNA in LLMs}. 
Just as DNA’s expression is shaped by regulatory logic rather than random variation, \textbf{nDNA encodes both universal priors (shared across tasks)} -- such as pretrained latent manifolds, attention mechanisms, and model architecture -- 
and \textbf{corpus-dependent latent scaffolding}, emerging as the model adapts to specific tasks or domains \citep{olah2020zoom, geva2021transformer, beltagy2020longformer}. 
The analogy emphasizes that corpus dependence in nDNA is not a weakness or artifact, but a reflection of meaningful task adaptation: 
\emph{structured variation grounded in universal latent geometry}. 
This scaffolding ensures LLMs achieve \textit{functional diversity} across tasks while maintaining \textbf{coherence, alignment, and generalization}, much like gene regulatory networks ensure appropriate cellular identity 
and function despite operating from a common genome blueprint \citep{alon2006introduction, davidson2006gene}.  
The figure highlights that both biological DNA and nDNA exhibit clarity through complexity: 
\textbf{layered, interpretable hierarchies enabling flexible, robust expression across contexts}.
}
\label{fig:enhanced_dna_ndna}
\vspace{-3mm}
\end{figure}
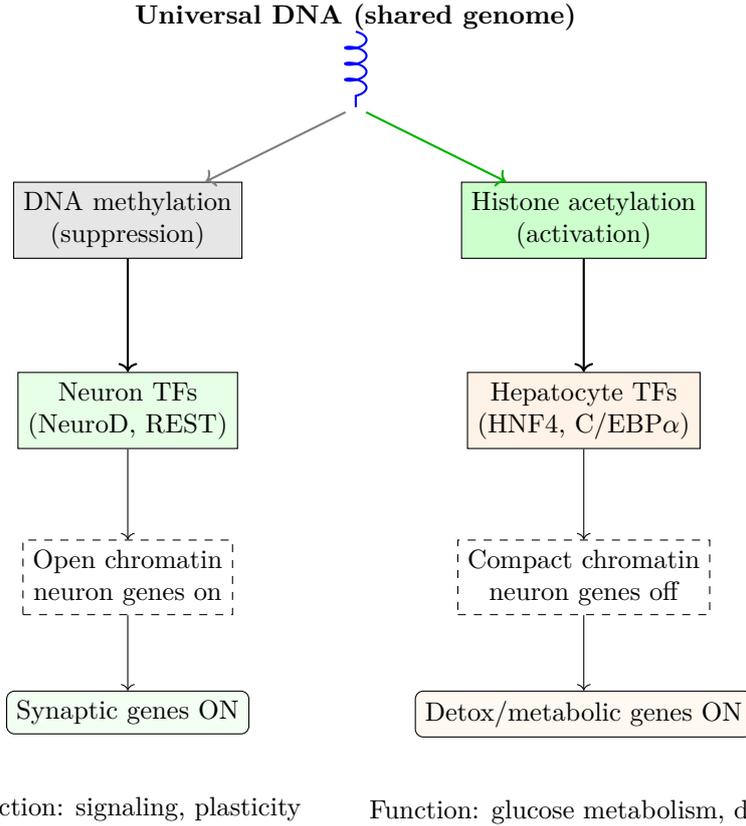

In biological systems, DNA is celebrated as the \emph{universal code of life} -- a sequence of nucleotides that, across all known organisms, governs the development, function, and inheritance of traits \citep{alberts2014molecular, lewin2013genes}. 
Yet despite this \textbf{universal structure}, the functional expression of DNA is profoundly \textbf{context-dependent}. The same genome, when expressed in different cellular contexts, gives rise to vastly different phenotypes: for instance, \textit{neurons} and \textit{hepatocytes} arise from identical genetic material yet serve radically different functions \citep{bird2007perceptions, davidson2006gene}. 
This context-sensitive expression is orchestrated through layered regulatory mechanisms, including \textbf{epigenetic modifications} \citep{bird2007perceptions}, \textbf{transcription factor (TF) binding} \citep{lambert2018human}, and \textbf{chromatin architecture remodeling} \citep{clapier2017mechanisms, dekker2013exploring}. 
These mechanisms form a hierarchical, probabilistic regulatory network that determines gene expression patterns in response to developmental and environmental cues \citep{alon2006introduction}. Figure \ref{fig:enhanced_dna_ndna} illustrates a hierarchical regulatory framework where universal DNA undergoes epigenetic modifications and context-specific transcription factor actions to produce specialized gene expression programs. Analogously, in large language models, this layered structure parallels \textbf{\textit{n}}DNA latent scaffolding that encodes both \emph{universal priors} and \emph{task-dependent adaptations}, enabling \textbf{coherent, flexible, and robust functional diversity} across domains.

Similarly, in large foundation models, the \emph{neural DNA (nDNA)} -- a composite measure of latent geometry encompassing \textbf{spectral curvature} ($\kappa$) \citep{belkin2019reconciling}, 
\textbf{thermodynamic length} ($L$) \citep{still2012thermodynamic}, 
and \textbf{latent belief vector norms} \citep{olah2020zoom} -- exhibits both \textbf{universal structure} and \textbf{corpus-specific adaptation}. 
LLMs encode universal latent priors through pretraining: architectural invariances \citep{vaswani2017attention}, semantic manifolds \citep{mikolov2013distributed, bommasani2021opportunities}, and attention-based relational structures \citep{geva2021transformer}. 
However, when probed with different corpora -- such as mathematical reasoning benchmarks (e.g. GSM8K \citep{cobbe2021training}), dialogue datasets (e.g. MultiWOZ \citep{budzianowski2018multiwoz}), or encyclopedic QA (e.g. SQuAD \citep{rajpurkar2016squad}) -- 
the model activates distinct latent scaffolding, producing task-specific geometric pathways.

In both systems, \textbf{structured variation emerges as a necessity}: 
in \textbf{biology}, to produce \emph{functional diversity} across cell types; 
in \textbf{LLMs}, to scaffold \emph{reasoning} across tasks while maintaining \textbf{alignment} and \textbf{generalization} \citep{bommasani2021opportunities, cobbe2021training}. 
Like \textbf{tissue-specific gene expression}, \textbf{corpus-dependent nDNA scaffolding} follows precise, \emph{learned priors} rather than arbitrary variation. 
\textbf{Mathematical models} of both systems reduce to \emph{path integrals over conditional cost}: 
\[
\mathcal{S}(c) = \int_{\gamma_c} \mathcal{C}(h_\ell; c) ds
\]
where $\gamma_c$ is the pathway for \emph{context} $c$ (cell type or corpus), and $\mathcal{C}$ reflects \emph{regulatory} or \emph{loss cost}.

\vspace{-2mm}
\begin{quote}
\emph{\textbf{Where DNA differentiates cells}, nDNA differentiates reasoning. Both systems achieve \textbf{functional coherence} through context-dependent geometry anchored in \textbf{universal code}.}
\end{quote}

Despite their \emph{contextual variation}, both \textbf{DNA} and \textbf{nDNA} encode \textbf{universal structure} that stabilizes functional diversity. 
In \textbf{biology}, this universality is embodied in the \emph{genetic code}: the shared language of \textbf{codons}, \textbf{conserved regulatory motifs}, and \textbf{chromatin architectural principles} that ensure coherent development across tissues \citep{lewin2013genes, alberts2014molecular}. 
In \textbf{large language models}, nDNA’s universality arises from the \textbf{shared latent priors} learned during pretraining: \textbf{attention-based relational structures} \citep{vaswani2017attention}, \textbf{semantic manifolds} \citep{mikolov2013distributed}, and \textbf{transformer-invariant latent symmetries} \citep{bommasani2021opportunities}. 
These priors act as the \emph{``genomic grammar''} that binds task-specific latent pathways into a \textbf{coherent reasoning framework}.

\[
\boxed{
\textbf{DNA: } \Sigma^3 / \ker \phi \to \mathcal{A} \quad 
\textbf{nDNA: } \mathcal{X} / G_{\mathrm{LLM}} \to V/G
}
\]

Such \textbf{universal structure} enables \textbf{generalization}: 
in \textbf{biology}, reliable \emph{organismal development}; 
in \textbf{LLMs}, reasoning \emph{consistency} and \emph{alignment} across tasks. 
Crucially, this structure constrains \textbf{corpus-dependent variation} within \emph{interpretable latent geometry} -- preventing arbitrary or adversarial drift \citep{geva2021transformer, cobbe2021training}. 

\vspace{-2mm}
\begin{quote}
\emph{\textbf{What DNA is to the unity of multicellular life}, nDNA is to the coherence of LLM reasoning: a \textbf{stabilizing universal code} that enables \textbf{structured functional variation}.}
\end{quote}

\subsection*{Evolutionary and learning dynamics: convergence of principles}
Both \textbf{DNA} and \textbf{nDNA} are shaped by \emph{selection processes}. 
In \textbf{biology}, the genome has evolved under millennia of selective pressure, with \textbf{regulatory networks} fine-tuned to ensure \emph{robust development} and \emph{adaptability} \citep{alon2006introduction, davidson2006gene}. 
In \textbf{LLMs}, pretraining operates as an \emph{evolutionary analogue}: \textbf{stochastic gradient descent (SGD)} over massive corpora selects latent priors that minimize expected loss across tasks, with \emph{fine-tuning akin to epigenetic adjustment} \citep{bommasani2021opportunities, pfeiffer2021adapterfusion}. 

\[
\underbrace{
\mathcal{L}_{\mathrm{pretrain}}(\theta) = \mathbb{E}_{(x,y)} \left[ -\log p_\theta(y|x) \right]
}_{\textbf{SGD as selection pressure}}
\]

This \textbf{evolutionary parallel} explains why both systems exhibit \emph{clarity through complexity}: \textbf{layered hierarchies}, \textbf{probabilistic pathways}, and \textbf{interpretable modularity}. 
Where \textbf{biological evolution} yields \emph{modular gene regulatory networks} that ensure context-sensitive expression \citep{alon2006introduction}, \textbf{LLM training} yields \emph{modular latent structures} -- such as \textbf{attention heads} and \textbf{adapter modules} -- that scaffold \emph{task-specific reasoning} \citep{geva2021transformer, pfeiffer2021adapterfusion}.

\subsection*{Why corpus dependence matters}
\begin{figure*}[t]
  \centering
  \begin{minipage}[t]{0.48\textwidth}
    \centering
    \includegraphics[width=\textwidth]{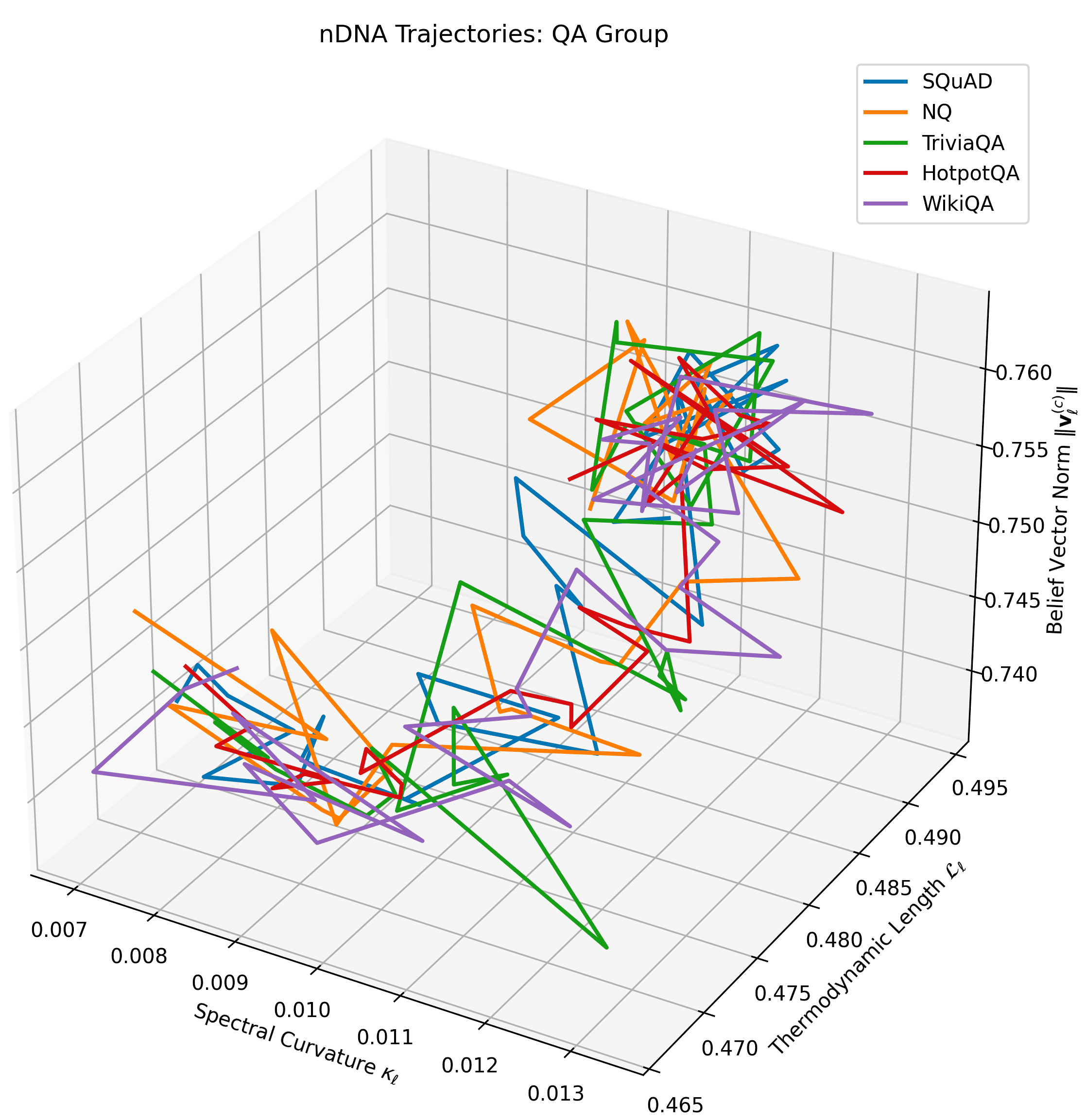}
    \subcaption{\textbf{QA group nDNA trajectories}: $\kappa$ ranges $\sim 0.012$--$0.03$, $L$ $\sim 0.47$--$0.53$, $\tau$ $\sim 0.006$--$0.014$. Trajectories are compact and consistently shaped across datasets, reflecting \textbf{shared task structure}.}
  \end{minipage}
  \hfill
  \begin{minipage}[t]{0.48\textwidth}
    \centering
    \includegraphics[width=\textwidth]{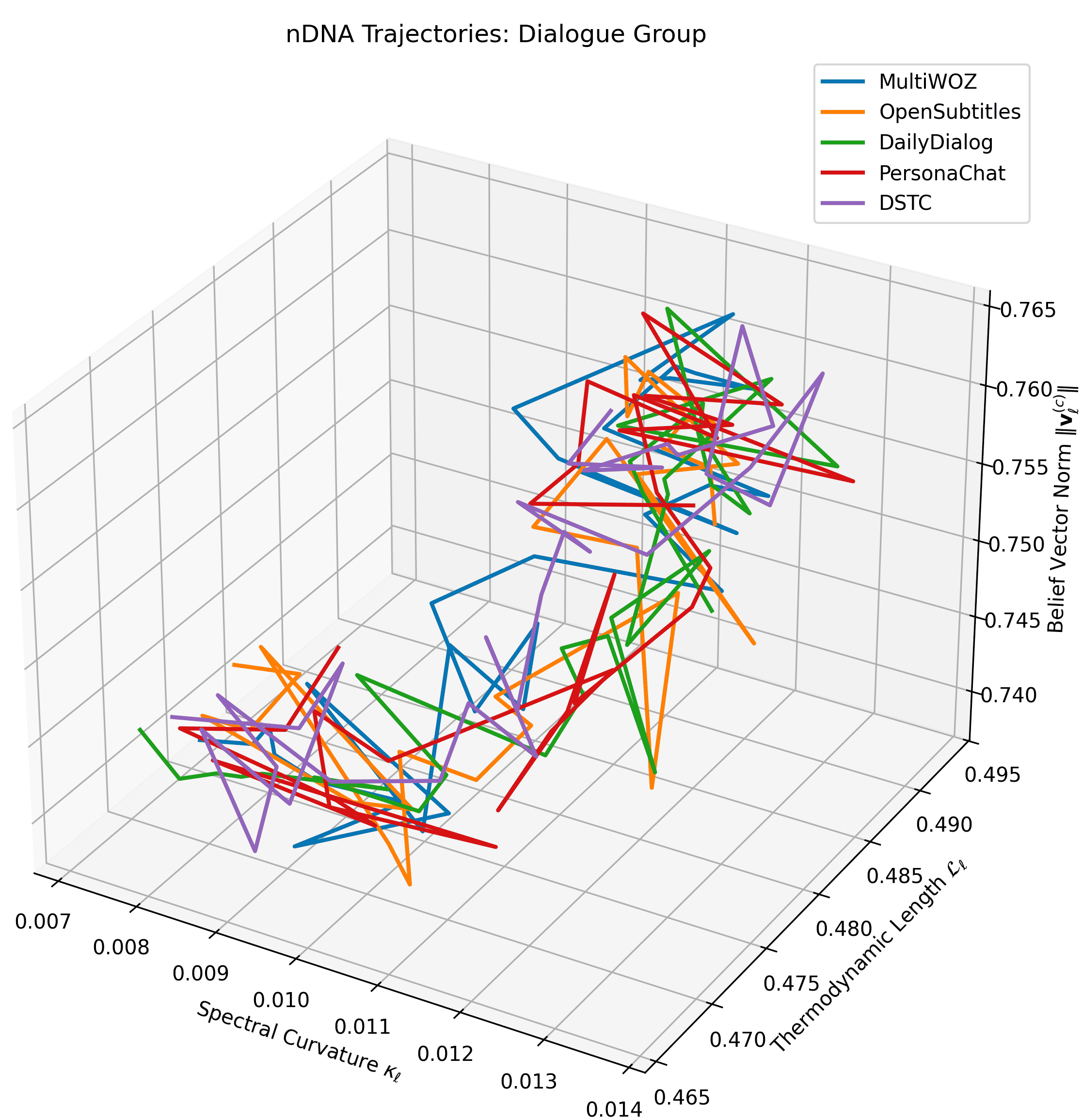}
    \subcaption{\textbf{Dialogue group nDNA trajectories}: $\kappa$ ranges $\sim 0.01$--$0.03$, $L$ $\sim 0.47$--$0.53$, $\tau$ $\sim 0.006$--$0.014$. Trajectories are shallow and tightly clustered, reflecting \textbf{low latent complexity} typical of conversational flow.}
  \end{minipage}
  \vspace{0.4cm}
  \begin{minipage}[t]{0.48\textwidth}
    \centering
    \includegraphics[width=\textwidth]{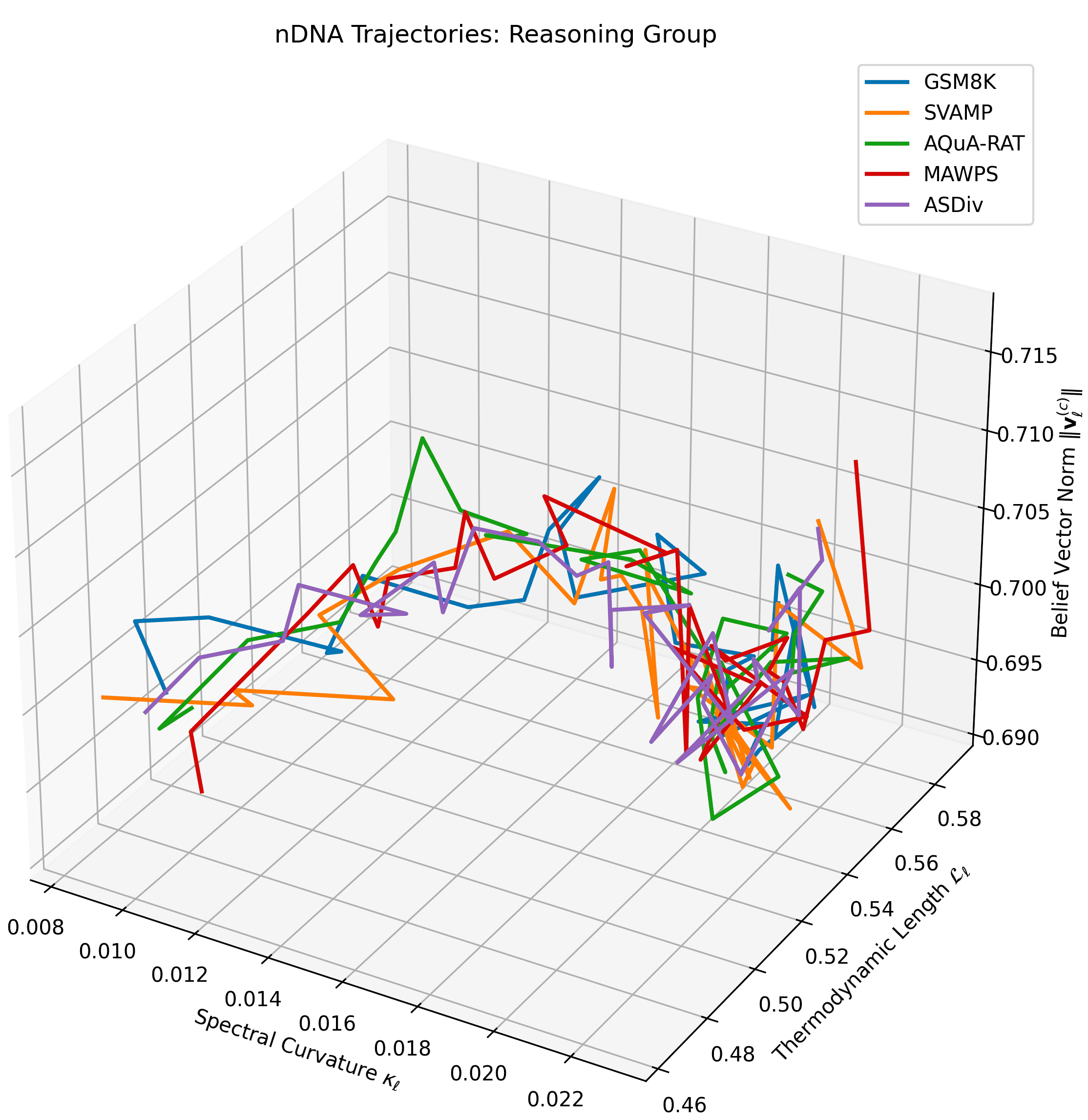}
    \subcaption{\textbf{Reasoning group nDNA trajectories}: $\kappa$ ranges $\sim 0.005$--$0.04$, $L$ $\sim 0.44$--$0.56$, $\tau$ $\sim 0.002$--$0.018$. Trajectories show \textbf{greater spread and complexity}, reflecting multi-step reasoning scaffolding.}
  \end{minipage}
  \hfill
  \begin{minipage}[t]{0.48\textwidth}
    \centering
    \includegraphics[width=\textwidth]{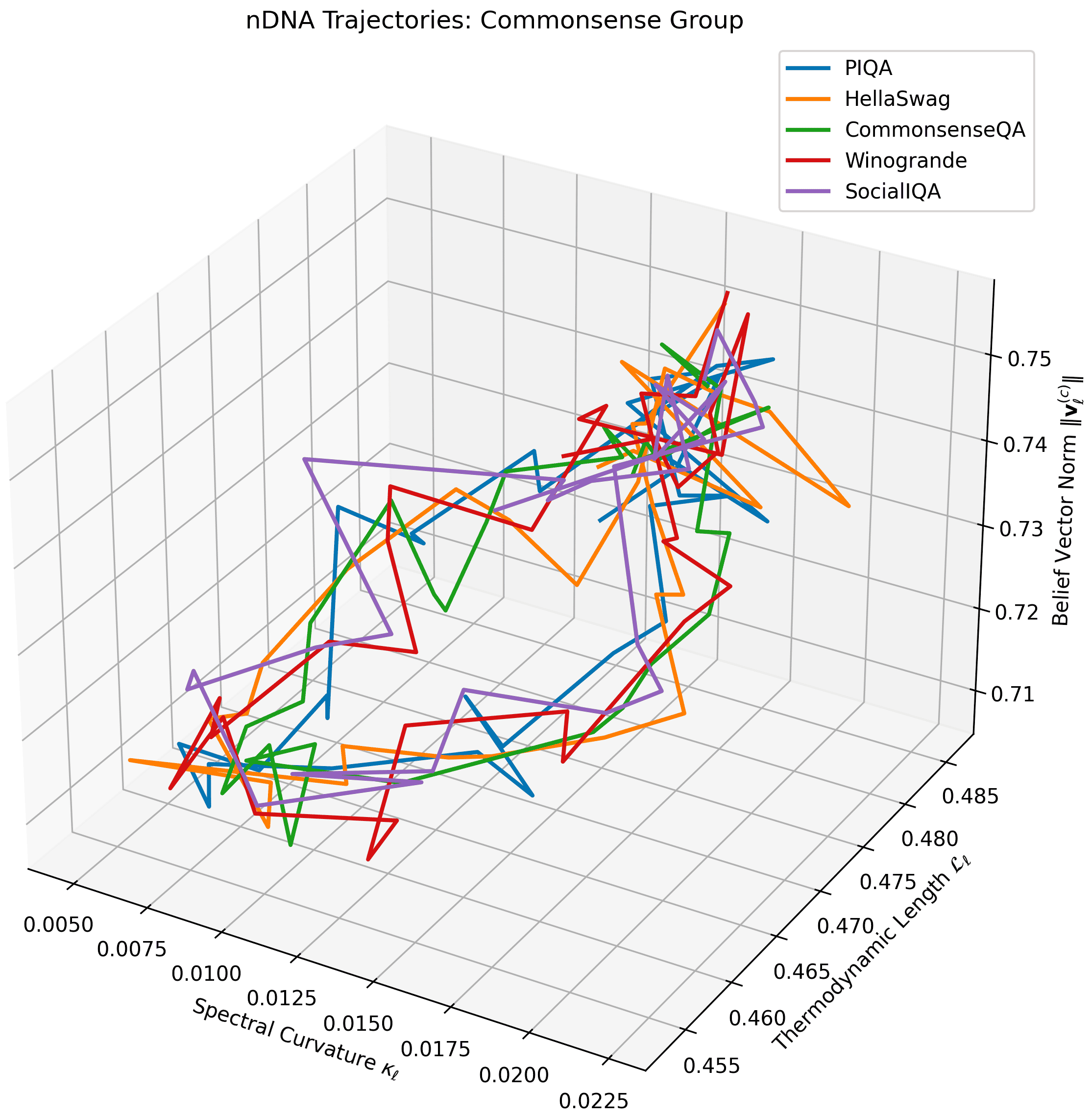}
    \subcaption{\textbf{Commonsense group nDNA trajectories}: $\kappa$ ranges $\sim 0.00$--$0.04$, $L$ $\sim 0.44$--$0.54$, $\tau$ $\sim 0.004$--$0.018$. Trajectories are intermediate in complexity, reflecting \textbf{varied latent demands of commonsense reasoning}.}
  \end{minipage}
  \vspace{-6mm}
  \caption{
    \textbf{nDNA trajectories across LLaMA vs. task groups.} 
    Each subplot visualizes \textbf{spectral curvature} ($\kappa_\ell$), \textbf{thermodynamic length} ($\mathcal{L}_\ell$), and \textbf{belief vector norm} ($\|\mathbf{v}_\ell^{(c)}\|$) layer-wise trajectories for representative datasets. 
    The structured variation illustrates that \emph{corpus dependence in nDNA is meaningful and interpretable}, reflecting task complexity rather than random noise. 
    \textbf{QA} and \textbf{dialogue} tasks activate \textbf{compact, smooth latent scaffolds with low curvature and modest belief steering}; \textbf{reasoning} tasks exhibit broader, more intricate geometry, with \textbf{increasing curvature, longer latent length, and stronger belief vector dynamics}. 
    \textbf{Commonsense} tasks show intermediate complexity with \textbf{oscillatory scaffolding, reflecting ambiguity and contextual switching}. 
    This figure demonstrates the core takeaway of our section: 
    \emph{like biological DNA, nDNA expresses differently in context, but remains bound by universal latent priors that ensure coherence, generalization, and alignment.}
    }
  \label{fig:ndna_groups}
\end{figure*}

Far from a flaw, \textbf{corpus dependence in \emph{n}DNA} is the signature of a \emph{flexible}, \emph{adaptive reasoning architecture}. 
Just as biological systems rely on \textbf{tissue-specific gene expression} to produce functional diversity from a \emph{universal genome} \citep{davidson2006gene, alon2006introduction}, large language models (LLMs) leverage \textbf{corpus-dependent latent scaffolding} to generate reasoning structures attuned to task demands, mirroring the reproducibility logic of biological variability quantification \citep{marioni2011rna}. 
By examining nDNA’s \textbf{spectral curvature} ($\kappa$), \textbf{thermodynamic length} ($\mathcal{L}$), and \textbf{belief vector norm} ($\|\mathbf{v}_\ell^{(c)}\|$), we gain a \textbf{diagnostic lens} for alignment, generalization, and safety \citep{belkin2019reconciling, still2012thermodynamic, olah2020zoom}:
\[
\mathcal{S}_{\mathrm{nDNA}}(c) = \int_{\gamma_c} \left( \alpha \kappa + \beta \mathcal{L} + \gamma \|\mathbf{v}_\ell^{(c)}\| \right) ds
\]
where $\gamma_c$ is the latent trajectory for corpus $c$. This latent geometry echoes Waddington’s epigenetic landscape where paths represent developmental fates \citep{waddington1957strategy}. 
Figure ~\ref{fig:ndna_groups} -- \textbf{QA tasks} evoke compact low-curvature paths (e.g.\ $\kappa \sim 0.012$--$0.03$, $\mathcal{L} \sim 0.47$--$0.53$) \citep{rajpurkar2016squad, kwiatkowski2019natural, joshi2017triviaqa}, while \textbf{reasoning tasks} elicit broader high-curvature paths (e.g.\ $\kappa \sim 0.005$--$0.04$) \citep{cobbe2021training, patel2021nlp, geva2021transformer}. 
\textbf{Dialogue corpora} produce shallow clustered scaffolds \citep{budzianowski2018multiwoz, li2016persona, zhang2018personalizing}; \textbf{commonsense tasks} yield oscillatory paths \citep{sap2019socialiqa, zellers2019hellaswag, talmor2019commonsenseqa}. 
nDNA aligns with interpretable AI goals \citep{zhang2018interpretable} and geometric decoding approaches \citep{narayanan2021decoding}.

This corpus dependence is \emph{not arbitrary noise} -- it reflects the model’s \textbf{learned latent regulatory logic}, analogous to the combinatorial control of \textbf{gene regulatory networks} that ensures \emph{context-sensitive yet robust gene expression} \citep{alon2006introduction, lewin2013genes}. 
Just as \emph{developmental disorders} arise when regulatory circuits misfire \citep{davidson2006gene}, misalignment or hallucination in LLMs can be traced to \emph{latent trajectories that diverge from expected scaffolding}. 
\textbf{nDNA analysis}, therefore, does not merely characterize model geometry -- it offers a \textbf{tool for interpretability, failure detection, and safe alignment}.

\vspace{-1mm}
\begin{quote}
\emph{\textbf{Corpus dependence in nDNA is the expression of reasoning plasticity}, bounded by universal latent priors much like gene networks balance flexibility with functional coherence.}
\end{quote}
\vspace{-1mm}

Moreover, the \textbf{universality of nDNA’s foundational structure} -- its pretrained manifold, architectural symmetries, and core alignment priors -- provides the \emph{stabilizing grammar} that constrains corpus-specific scaffolds within meaningful reasoning spaces \citep{vaswani2017attention, bommasani2021opportunities}. 
This is the latent equivalent of biology’s \textbf{genetic code} and \textbf{conserved transcriptional machinery}: an \emph{invariant substrate} that supports functional diversity without sacrificing coherence. 
By quantifying how nDNA paths \emph{bend}, \emph{stretch}, or \emph{steer} in response to task demands, we can map the model’s \textbf{cognitive landscape} -- and determine when it traces \emph{human-aligned reasoning} or drifts into failure modes.

\vspace{-1mm}
\begin{quote}
\emph{\textbf{What the genome is to life's functional unity}, nDNA is to the model’s reasoning coherence: a universal code that binds diversity into stability, and complexity into interpretability.}
\end{quote}
\vspace{-1mm}

\begin{table}[H]
\centering
\caption*{\centering \textbf{Mathematical comparison of DNA and nDNA structural layers}}
\begin{tabular}{p{3cm} | p{5.6cm} | p{5.6cm}}
\toprule
\textbf{Layer} & \textbf{DNA (Biology)} & \textbf{nDNA (LLM)} \\
\midrule
\textbf{Universal code} & Codon mapping $\phi: \Sigma^3 \to \mathcal{A}$, kernel $\neq \emptyset$, redundancy ensures error tolerance \citep{lewin2013genes} & Pretrained latent manifold; symmetries $G_{\mathrm{LLM}} \subset \mathrm{Aut}(V)$; generalization via equivariance \citep{bommasani2021opportunities} \\
\textbf{Context regulator} & Conditional $P(\text{gene ON}|\text{TF, epi})$; Bayesian gene networks \citep{alon2006introduction} & Conditional latent path $P(h_1,\dots,h_L|x)$; stochastic latent dynamics \citep{geva2021transformer} \\
\textbf{Path geometry} & Minimal energy path $\gamma^*$ in epigenetic landscape: $\int_\gamma \|\nabla V\| ds$ \citep{waddington1957strategy} & Latent geodesic minimizing cost: $\int_\gamma \|\nabla_\theta \log p(y|x)\|^2 ds$ \citep{still2012thermodynamic} \\
\textbf{Output mapping} & Fiber bundle: $\pi: E_{\mathrm{gene}} \to B_{\mathrm{cell}}$ & Fiber bundle: $\pi: E_{\mathrm{latent}} \to B_{\mathrm{task}}$ \\
\bottomrule
\end{tabular}
\end{table}

\clearpage
\newpage

\clearpage
\newpage






\end{document}